%% file: main.tex
\pdfoutput=1
\documentclass[10pt,twocolumn,letterpaper]{article}

\usepackage[pagenumbers]{cvpr} %

\input{preamble}

\definecolor{cvprblue}{rgb}{0.21,0.49,0.74}
\usepackage[pagebackref,breaklinks,colorlinks,allcolors=cvprblue]{hyperref}

\newcommand{\insertfig}{%
  \captionsetup{type=figure}%
  \includegraphics[width=\textwidth]{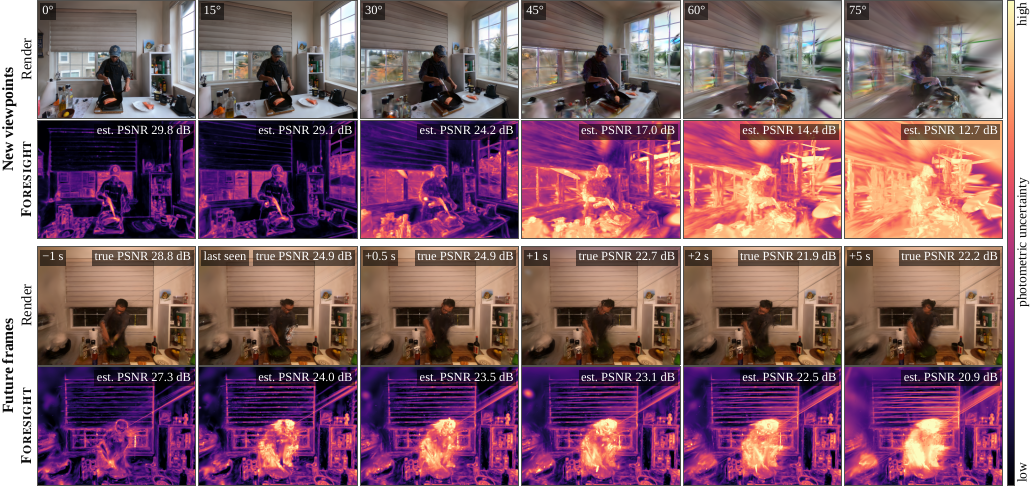}%
  \vspace{-6pt}
  \caption{
    \ours estimates the \emph{per-pixel uncertainty} of a frozen dynamic Gaussian splatting model's renders at any
    viewpoint and time. Top: a virtual viewpoint rotated away from a test camera
    (N3DV~\cite{li2022neural}, 4DGaussians~\cite{wu2024_4dgaussians}). Bottom: future frames
    forecast by EvoGS~\cite{asiimwe2025evogs}. Renders are labeled with their true PSNR
    where ground truth exists, maps with the PSNR that \ours estimates without it. Uncertainty grows where rendering
    degrades: across the view as the viewpoint leaves the rig, and on the moving person as the horizon grows.}%
  \label{fig:teaser}}
\makeatletter
\apptocmd{\@maketitle}{\vspace{-12pt}\centering\insertfig\vspace{20pt}}{}{}
\patchcmd{\@maketitle}{\vskip .375in}{}{}{}%
\makeatother

\def\confName{CVPR}
\def\confYear{2027}

\title{\textsc{Foresight}: Post-hoc Predictive Uncertainty for Dynamic Gaussian Splatting\\at Novel Views and Times}

\author{%
Chamuditha Jayanga Galappaththige$^{1,2*}$ \quad Peter Stehr$^{3*}$ \quad Thomas Gottwald$^{3}$ \quad Edgar Heinert$^{4}$\\
Niko Suenderhauf$^{1,2}$ \quad Matthias Rottmann$^{4}$ \quad Dimity Miller$^{1,2}$\\[0.4em]
{\small $^1$QUT Centre for Robotics \quad $^2$ARIAM Hub \quad $^3$University of Wuppertal \quad $^4$University of Osnabr\"uck}\\
{\small $^*$Equal contribution}
}

\begin{document}
\maketitle
\input{sec/00_abstract}    
\input{sec/01_intro}
\input{sec/02_related_work}
\input{sec/03_method}
\input{sec/04_experiments}
\input{sec/05_limitations}
\input{sec/06_conclusion}
{
    \small
    \bibliographystyle{ieeenat_fullname}
    \bibliography{main}
}

\setcounter{dbltopnumber}{3}\renewcommand{\dbltopfraction}{0.9}  %
\input{sec/XX_suppl}

\end{document}

%% file: preamble.tex
\newcommand{\redmath}{\color{black}}

\newcommand{\bluemath}{\color{black}}

\DeclareRobustCommand{\fstep}[1]{\tikz[baseline=(fs.base)]\node[circle,draw,line width=0.4pt,inner sep=0.3pt,minimum size=1.1em,font=\scriptsize\bfseries](fs){#1};}

\usepackage{microtype}
\usepackage{comment}
\makeatletter
\renewcommand\paragraph{\@startsection{paragraph}{4}{\z@}{1.5ex \@plus .1ex \@minus .1ex}{-1em}{\normalfont\normalsize\bfseries}}
\makeatother

\usepackage{tikz}
\usetikzlibrary{arrows.meta,calc}

\usepackage{booktabs}
\usepackage{colortbl}
\usepackage[bb=boondox]{mathalfa}
\definecolor{tabfirst}{RGB}{255,201,153}   %
\definecolor{tabsecond}{RGB}{255,236,204}  %
\newcommand{\first}[1]{\cellcolor{tabfirst}\textbf{#1}}
\newcommand{\second}[1]{\cellcolor{tabsecond}#1}
\DeclareRobustCommand{\firstbox}[1]{{\setlength{\fboxsep}{1pt}\colorbox{tabfirst}{\textbf{#1}}}}
\DeclareRobustCommand{\secondbox}[1]{{\setlength{\fboxsep}{1pt}\colorbox{tabsecond}{#1}}}

\newcommand{\ours}{\mbox{\textsc{Foresight}}\xspace}  %
\newcommand{\vbeta}{\boldsymbol{\beta}}
\newcommand{\vtheta}{\boldsymbol{\theta}}
\newcommand{\vpsi}{\boldsymbol{\psi}}
\newcommand{\vmu}{\boldsymbol{\mu}}
\newcommand{\vnu}{\boldsymbol{\nu}}

\newcommand{\vomega}{\boldsymbol{\omega}}

\newcommand{\mSigma}{\boldsymbol{\Sigma}}
\newcommand{\mPi}{\boldsymbol{\Pi}}
\newcommand{\const}{C}
\DeclareMathOperator{\blockdiag}{blockdiag}
\DeclareMathOperator{\diag}{diag}

%% file: sec/00_abstract.tex
\begin{abstract}
Dynamic Gaussian splatting renders a captured event from any viewpoint at any instant, enabling free-viewpoint replay
and volumetric video.
Its render quality varies by several decibels across viewpoints and time, yet at novel viewpoints and times no
ground truth is available to assess it, so applications must estimate where, and by how much, a render is wrong
before showing it.
Post-hoc estimators of the rendering error exist for static scenes, and dynamic methods use uncertainty during
training to improve the reconstruction; neither estimates the error of an already trained dynamic model.
We therefore present \textup{\ours}, a post-hoc method that estimates the per-pixel variance of the rendering
error of an already trained dynamic Gaussian splatting model at any viewpoint and time, even beyond the captured span.
Inspired by regression prediction intervals, it multiplies the error observed on the training views by one plus a
\emph{leverage}, a measure of how poorly those views constrain a novel render, and calibrates the result on one
held-out camera.
In five settings, from dense and sparse rigs to monocular capture and future frames, each with a different backbone
family, \textup{\ours} localizes the error with 11--35\% lower AUSE than the best of three post-hoc methods for static
scenes adapted to these backbones, with a better likelihood (NLL).
Because the variance is calibrated, its mean over a view estimates the view's PSNR with a mean absolute error of
0.61--2.03\,dB.
Project page: \href{https://chumsy0725.github.io/foresight}{github.io/foresight}
\end{abstract}

%% file: sec/01_intro.tex
\section{Introduction}
\label{sec:intro}
\begin{figure*}[t]
  \centering
  \includegraphics[width=\linewidth]{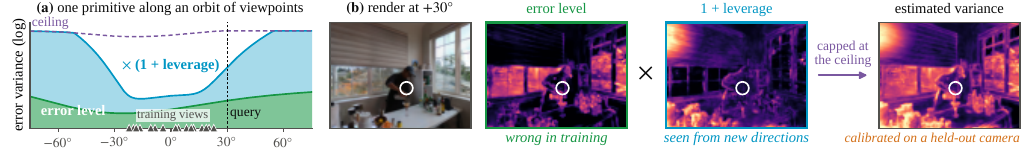}
  \vspace{-2em}
  \caption{
    \ours estimates the variance of a render's error as in a regression prediction interval: the training residuals
    alone underestimate the error away from the training views, and the leverage widens the estimate there.
    (a) One primitive as the viewpoint orbits the scene (N3DV, 4DGaussians; log scale): its error level, fitted to
    the training residuals (green), times one plus its leverage (cyan), low near the training views
    (\textcolor{black!70}{$\blacktriangle$}) and growing away from them, up to the ceiling (purple).
    (b) The same terms per pixel at a viewpoint 30$^\circ$ outside the rig (no ground truth), each pixel
    mixing the leverages of its primitives; $\circ$~marks the primitive of (a). Their product, capped at the ceiling
    and calibrated on a held-out camera, is the estimate. Error level and estimate share one log color scale;
    black leverage means no increase.}
  \label{fig:idea}
\end{figure*}

3D Gaussian Splatting (3DGS)~\cite{kerbl2023gaussian} renders photorealistic novel views in real time, and recent
works extend it to dynamic scenes~\cite{wu2026atgs,li2024spacetime,duan20244d,wu2024_4dgaussians,%
yang2024deformable,yang2024real,zhou20264c4d,wang2025shape,lei2025mosca}, so that a captured event can be replayed
from any viewpoint at any moment, \eg in free-viewpoint sports replays~\cite{wu2026atgs} and volumetric
video~\cite{wang2024v3,jiang2024dualgs}, or even
forecast beyond the end of the capture~\cite{asiimwe2025evogs}.
With few cameras per instant and surfaces that move and occlude each other, render quality varies by several
decibels across viewpoints and time (\cref{fig:teaser}).
Novel viewpoints and future times were never recorded, so no ground truth reveals where a render fails, and checking
every view and frame by eye does not scale.
Free-viewpoint video therefore needs an automatic estimate of where, and by how much, each render is wrong, so that
failed views can be avoided.

For static scenes, uncertainty estimation relies on stochastic
training~\cite{shen2021stochastic,aira2025modelinguncertainty,li2024variational,lyu2024manifold}, Fisher information
of the primitives' parameters~\cite{jiang2024fisherrf,wilson2025popgs,hanson2025pup}, or post-hoc estimators of the rendering
error~\cite{galappaththige2026predictive,gottwald2026primu}.
In dynamic Gaussian splatting, uncertainty serves to improve the reconstruction itself: as a training signal~\cite{kim20244d,guo2026uncertainty,kim2026gp}, or to select
training views by extending FisherRF~\cite{jiang2024fisherrf} to dynamic scenes~\cite{li2026next}.
None of these dynamic methods estimate the predictive uncertainty of an already trained model, and most are specific to
a particular model architecture.

We present \ours to answer this need.
Given a trained dynamic Gaussian splatting model and its training images, it estimates post hoc the
predictive uncertainty of the render at any viewpoint and time, even beyond the captured span, as the per-pixel
variance of its rendering error (\cref{fig:teaser}).
Training residuals show where the model fails on the training views, but underestimate the error at new
viewpoints and times.
As in a regression prediction interval, which widens the farther a new input lies from the training inputs, \ours
therefore multiplies an error level by one plus a leverage (\cref{fig:idea}).
First, it fits the \emph{error level} to the training residuals, per primitive over viewing direction and time, and
splats it to the pixels like a color.
Second, it computes the \emph{leverage}, which measures how poorly the training views, weighted by this error
level, constrain each primitive's appearance at the queried direction and time.
The leverage is high for primitives rendered from new directions or times.
Third, beyond the captured time span, it adds a \emph{time} term based on the changes in the backbone's own renders.
Fourth, it caps the estimate at a \emph{ceiling}, the error of an uninformed guess from the training colors, and moves
pixels with little reconstructed geometry toward this ceiling.
Finally, four scalars fitted on one held-out camera, used for neither training nor testing, calibrate the variance,
so that its mean over a view estimates the PSNR.

\noindent In summary, our contributions are:
\begin{itemize}
    \item To our knowledge, \ours is the first post-hoc estimator of the predictive uncertainty of dynamic Gaussian
    splatting: a per-pixel variance of the rendering error at any viewpoint and time that uses the backbone only
    through its renders and rasterizer (\cref{sec:method}).
    \item \ours derives this variance as in a regression prediction interval: an \emph{error level} fitted to the
    training residuals, times one plus a \emph{leverage} that grows away from the training views, and both come from one linear
    model of the rendered colors (\cref{sec:method}).
    \item An extensive evaluation and analysis on dense and sparse rigs, monocular capture and future frames, with
    five backbone families: \ours has 11--35\% lower AUSE and a better NLL than the best of three post-hoc methods for static scenes adapted to these
    backbones; calibrated on one held-out camera, it estimates a view's PSNR with a mean absolute error of
    0.61--2.03\,dB (\cref{sec:experiments}).
\end{itemize}

%% file: sec/02_related_work.tex
\section{Related Work}
\label{sec:related_work}

\noindent\textbf{Dynamic Gaussian splatting.}
3DGS~\cite{kerbl2023gaussian} has been extended to dynamic scenes by deforming canonical
primitives with a learned field~\cite{wu2024_4dgaussians,yang2024deformable}, by native 4D primitives sliced in
time~\cite{yang2024real,duan20244d}, by per-primitive temporal functions~\cite{li2024spacetime}, and by low-dimensional
motion bases or scaffolds for monocular video~\cite{wang2025shape,lei2025mosca}.
Recent methods target four portable cameras~\cite{zhou20264c4d} or long sequences~\cite{wu2026atgs}, or render
future frames using graph convolutional networks~\cite{zhao2024gaussianprediction}, latent
ODEs~\cite{wang2026odegs}, motion-aware groups of primitives~\cite{lee2026mogaf} or continuous-time dynamical
models~\cite{asiimwe2025evogs}, so far evaluated on synthetic scenes and handheld
captures~\cite{zhao2024gaussianprediction,wang2026odegs,lee2026mogaf,asiimwe2025evogs} and on multi-view
capture with moderate motion~\cite{asiimwe2025evogs}.
These methods report floaters and inconsistent motion~\cite{duan20244d}, difficulties with large
motion~\cite{wu2024_4dgaussians}, and jitter where viewpoints are very sparse~\cite{wu2026atgs}.
As such failures depend on the viewpoint and time, free-viewpoint video needs a per-pixel
estimate of the rendering error of a trained model at unrecorded viewpoints and times.

\noindent\textbf{Uncertainty in radiance fields.}
In Gaussian splatting, as earlier in neural radiance fields (NeRF)~\cite{mildenhall2020nerf}, uncertainty has mainly been estimated for two quantities, which we relate at the end of this section: the
parameters of the representation and the photometric error of its renderings.
For the parameters, training-time methods place a variational or sampled distribution over the
scene~\cite{li2024variational,aira2025modelinguncertainty,lyu2024manifold}, and post-hoc methods
derive it from a Laplace approximation~\cite{goli2024bayesrays}, Fisher information summed per primitive~\cite{jiang2024fisherrf}, or
per-primitive Fisher blocks~\cite{hanson2025pup,wilson2025popgs} of a trained model.
Their pixel maps show how well the data constrain these parameters,
as do maps of how visible each primitive is from the training views~\cite{xue2026uncertainty}.
For the photometric error, training-time methods learn a variance per sample~\cite{pan2022activenerf} or per
primitive~\cite{tran2026varsplat}, or evidential outputs~\cite{duan2026evidential}, jointly with a
static NeRF or 3DGS, so they apply only to the model trained with them.
Post hoc, \mbox{3DGS-U}~\cite{galappaththige2026predictive} fits the training residuals per
primitive over viewing direction to estimate the error at novel views, and PRIMU~\cite{gottwald2026primu} estimates it
from per-primitive error and coverage features with a learned regressor.
Both target static scenes and evaluate how well their maps correlate with the error.
Across these kinds, uncertainty often serves to prune primitives~\cite{hanson2025pup} or to choose what to capture
next~\cite{pan2022activenerf,jiang2024fisherrf,wilson2025popgs,xie2025gauss,ewen2026these,xue2026uncertainty,duan2026dynactivegs,galappaththige2026predictive,gottwald2026primu}.
In dynamic Gaussian splatting, uncertainty likewise serves the reconstruction and mostly concerns the parameters:
per-Gaussian coverage weights regularizers~\cite{kim20244d}, a time-varying variance from the curvature of the
photometric loss guides motion~\cite{guo2026uncertainty}, a Gaussian process models and extrapolates motion, with its
variance evaluated by sparsification on monocular scenes~\cite{kim2026gp}, and Fisher information selects viewpoints
to train on~\cite{li2026next}.
Others treat sparse views, missing frames, or unsynchronized cameras as uncertain inputs~\cite{yilmaz2026graphical},
or select viewpoints and times by the error and sensitivity of the rendering~\cite{otonari2026fillgs}.
These methods change how the model is trained, so they estimate the uncertainty of the model trained with them,
not of an already trained one.

The two kinds are complementary: photometric uncertainty measures how large the rendering error is where the
training views observe it, and parameter uncertainty shows where the data constrain the model poorly, thus,
where more views would help.
A regression prediction interval~\cite{hoaglin1978hat} relates the two: the variance of the error at a new query
is the noise variance, which the training residuals measure, times one plus the query's leverage, which grows the
farther the query lies from the training inputs.
For a render, these inputs are the directions and times at which the training views saw each primitive, so the
leverage must compare them with the query's.
Naively, one could multiply an error regression such as \mbox{3DGS-U} by one plus a parameter uncertainty such
as FisherRF's; on a sparse rig, this pairing performs like \mbox{3DGS-U} alone (\cref{tab:swap}), as FisherRF sums
each primitive's color information over its spherical-harmonic coefficients and the training views, so its score
does not depend on the query's direction or time.
\ours instead derives both terms from \emph{one linear color model} of the frozen backbone, whose leverage is that of a
correction of the rendered color over viewing direction and time, evaluated at the query, with the fitted error level
as noise.
It is this formulation of the leverage, not the pairing of the two kinds, that improves the estimate (\cref{sec:ablation}).

%% file: sec/03_method.tex
\section{Method}
\label{sec:method}

Given a frozen backbone and its training images $I_f$ at times $t_f\in[0,1]$, \ours estimates the expected squared
error of the render $\hat I$ at each pixel $p$ for any viewpoint and time,
\begin{equation}
  \sigma^2(p)\approx\mathbb E\bigl[\tfrac13\lVert\hat I(p)-I(p)\rVert^2\bigr],
  \label{eq:per-pixel-var}
\end{equation}
over the unknown true image $I$; for a zero-mean error, this is its variance.
We express the predictive uncertainty of the render by this per-pixel variance; its mean over a view estimates
the view's mean squared error and thus its PSNR.

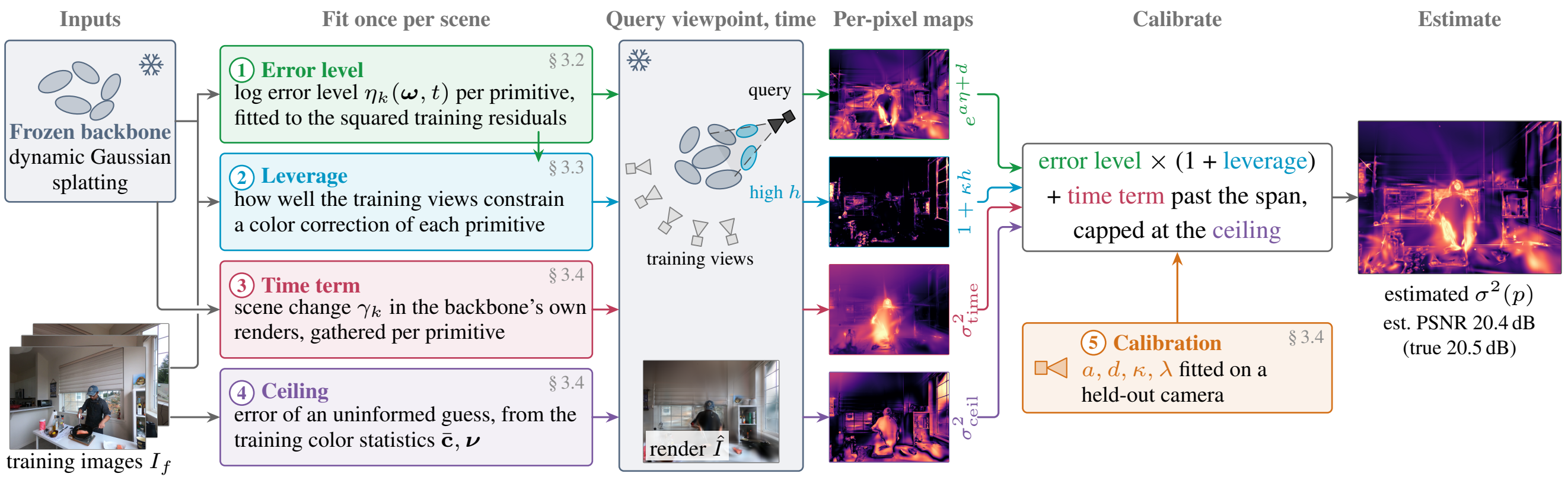
\begin{figure*}[t]
  \centering
  \input{figs/method/method.tex}
  \vspace{-0.7em}
  \caption{
    Overview of \ours.
    Once per scene, it fits the error level to the training residuals~\fstep{1}, measures how well the training views
    constrain a linear color model over direction and time, with this error level as noise~\fstep{2}, gathers the scene
    change in the backbone's own renders~\fstep{3}, and takes the training color statistics for the ceiling~\fstep{4};
    at a query viewpoint and time, the per-primitive terms are splatted and the ceiling is computed from the render,
    giving the estimate (\cref{eq:final}) with four scalars fitted on a held-out camera~\fstep{5}.}
  \label{fig:overview}
\end{figure*}

\ours derives this variance as in a regression prediction interval~\cite{hoaglin1978hat},
$\sigma^2=e^{\eta}(1+h)$ (\cref{fig:idea}): the \emph{error level} $e^{\eta}$ is the size of the errors at the
training views, and the \emph{leverage} $h$ widens the interval as the query moves away from them.
\Cref{sec:field,sec:leverage} derive both terms from a linear color model, and \cref{sec:open} adds a
\emph{time} term beyond the observed span, a \emph{ceiling}, and four scalars fitted on a held-out camera
(\cref{fig:overview}).

\subsection{Dynamic Gaussian splatting}
\label{sec:prelim}

At each time $t$, a dynamic backbone is a static 3DGS~\cite{kerbl2023gaussian} with primitives moved, deformed
or faded by its motion model.
A viewpoint is rendered by alpha-compositing them in depth order,
\begin{equation}
  \hat I(p) = \sum_k w_k(p)\,\mathbf c_k, \qquad
  w_k(p) = \alpha_k(p)\prod_{j<k}\bigl(1-\alpha_j(p)\bigr),
  \label{eq:composite}
\end{equation}
with $\alpha_k(p)$ the opacity of primitive $k$ at pixel $p$ and $\mathbf c_k$ its color seen from the viewpoint;
$A(p)=\sum_kw_k(p)\le1$ is the accumulated opacity.
With the geometry frozen, the weights are fixed, and the render is linear in per-primitive appearance
values~\cite{galappaththige2026predictive}.
Besides the backbone's own color render, \ours therefore uses the rasterizer for two operations only,
both available for any motion model: it \emph{splats} a per-primitive scalar $\phi_k$ in place of the color,
$\sum_kw_k(p)\,\phi_k$, which turns each per-primitive term below into a pixel map (\cref{fig:overview}), and \emph{gathers} a
pixel map $y$ onto the primitives with the backward pass of that render, $\sum_pw_k(p)\,y(p)$. 

\subsection{Error level}
\label{sec:field}

The training residuals are known only at the training views; the error level transfers them, as a variance, to
any viewpoint and time.
Each primitive carries its own error level, which varies with the viewing direction like a view-dependent color,
as in \mbox{3DGS-U}~\cite{galappaththige2026predictive}; we extend it over time.
Its logarithm is linear in features $\vpsi_k$ of direction and time, and the pixel's standard deviation is splatted
like a color:
\begin{equation}
\begin{gathered}
  \eta_k = \vpsi_k^\top\vbeta_k,\qquad \vpsi_k=\bigl(\mathbf Y(\vomega_k);\,\tilde{\mathbf B}(t)\bigr),\\
  \bluemath\eta(p) = 2\log\Bigl(\textstyle\sum_k w_k(p)\,e^{\eta_k/2} + \bigl(1-A(p)\bigr)\sqrt b\Bigr) + g(t).
\end{gathered}
\label{eq:eta}
\end{equation}
Here $\vbeta_k$ are the fitted coefficients, $\mathbf Y$ are the real spherical harmonics (SH) up to degree $L=3$
of the viewing direction $\vomega_k$ in the primitive's own frame where the backbone gives it one,
and $\tilde{\mathbf B}=\mathbf B-\frac1{12}\mathbf 1$ are 12 clamped cubic B-splines $\mathbf B$ in time, shifted
to sum to zero rather than one, so that a primitive's constant level sits in its SH part.
The uncovered part $1-A(p)$ of a pixel takes the prior level $b$ (below), while $g(t)$, a scene-global offset in
the same splines, shifts every pixel.

\textbf{Fitting.}
One pixel's squared residual $\frac13\lVert\hat I(p)-I(p)\rVert^2$ is a variance estimate with a spread of the
order of its mean, so we pool the squared training residuals of covered pixels ($A(p)>\frac12$) locally into targets
$y(p)$.
We fit $\eta$ with the loss of Gamma generalized linear models with log link~\cite{mccullagh1989generalized},
$\ell(\eta,y)=\eta+y\,e^{-\eta}$: its expectation is minimized at $e^{\eta}=\mathbb E[y]$ for any distribution
of $y$, and for a single squared residual it is twice the Gaussian negative log-likelihood per color channel, up to a constant, with
which we calibrate and evaluate.
Unlike the least-squares fit of $L_1$ and D-SSIM error in \mbox{3DGS-U}, this fits a variance that \cref{sec:leverage} takes as noise and penalizes
its relative misfit, as the targets span more than two orders of magnitude within a scene.

\textbf{Prior.}
The residuals sample a primitive's error level at only a few directions and times and leave it weakly determined at
all others.
A Gaussian smoothness prior with precision $\mPi$ therefore pulls each primitive toward a constant prior level
$b$, as the sparse-view prior of \mbox{3DGS-U} does.
We set $b$ to a high percentile of the targets, so an unobserved primitive is assumed to be as bad as the badly
reconstructed pixels.
It encodes our assumption that a primitive's error level varies smoothly with viewing direction and over time:
over directions, its covariance is a heat kernel on the sphere, which is diagonal in the SH and shrinks higher
degrees more strongly~\cite{borovitskiy2020matern}; over time, it is a Mat\'ern-3/2 covariance between the spline
coefficients~\cite{rasmussen2006gaussian}.
We minimize the mean loss over sampled training pixels plus the prior penalty with small weights, weaker on the
constant level, and a ridge on $g$ (weights and schedule in the supplement).

\subsection{Leverage}
\label{sec:leverage}

Training residuals are optimistic where the training views constrain the render poorly, \eg on surfaces seen from
new directions or at rarely observed times.
In linear regression, the leverage $\mathbf x^\top(\mathbf X^\top\mathbf X)^{-1}\mathbf x$ of a new input
$\mathbf x$ is small near many training inputs and grows away from them~\cite{hoaglin1978hat}, as the cyan band
in \cref{fig:idea}a shows for one primitive.
We build the same quantity for a pixel by asking how precisely the training images, with the error level of
\cref{sec:field} as their noise, could have pinned down a deviation in color at the query.

\textbf{Linear color model.}
We set up this regression on the colors alone: the geometry is taken as given and the render as the fitted value;
errors of geometry and motion that the training views reveal enter through the error level.
Suppose each color channel of primitive $k$ is off by its own unknown correction $\vpsi_k^\top\vtheta_k$ in the
basis of the error level (\cref{eq:eta}), so that pixel $p$ is off by $\mathbf x(p)^\top \vtheta = \sum_k w_k(p)\,\vpsi_k^\top\vtheta_k$ (\cref{eq:composite}), where $\vtheta$ stacks the $\vtheta_k$.
We model the residuals of each color channel as:
\begin{equation}
    r(p) = \mathbf x(p)^\top \vtheta + \epsilon(p),
    \label{eq:color-correction}
\end{equation}
with independent noise
$\epsilon(p) \sim \mathcal N\bigl(0,e^{\eta(p)}\bigr)$ at the fitted level, which we hold fixed: we infer only the
uncertainty of the correction, not the error level again. On $\vtheta_k$, we impose  the prior
$\vtheta_k\sim\mathcal N(\mathbf 0,b\,\mPi^{-1})$ built from the precision $\mPi$ and level $b$ of \cref{sec:field}:
the correction is as smooth as the error level and, where unobserved, of the order of $b$.
The correction is never estimated: its closed-form posterior covariance, the same for the three channels, depends on how often, from which directions and at which times each primitive was observed, weighted by the error level, and on the residuals only through this level.

\textbf{Fisher blocks.}
As the full posterior precision (derivation in the supplement) couples all primitives, we keep a
block per primitive, as PUP~3D-GS~\cite{hanson2025pup} and POp-GS~\cite{wilson2025popgs} do:
\begin{equation}
\begin{gathered}
    \mSigma_k=(\mathbf H_k+\mPi/b)^{-1}, \\
    \mathbf H_k=\textstyle\sum_{f=(c,t)}\bigl(\sum_pw_k(p;c,t)^2\,e^{-\eta(p;c,t)}\bigr)\vpsi_{kf}\vpsi_{kf}^\top,
\end{gathered}
\end{equation}
where the Fisher block $\mathbf H_k$ counts the training pixels that saw primitive $k$, each by its squared
weight over its error level, at the direction and time $\vpsi_{kf}$ of training images $f$, and $\mPi/b$ comes from
the prior.
In contrast to PUP~3D-GS and POp-GS, whose blocks are over the primitives' own parameters at constant noise, ours are over the color correction in the
basis $\vpsi_k$ of direction and time. \Cref{sec:ablation} compares this leverage with one built from FisherRF's
per-primitive information on a sparse rig.
For a query pixel $q$, we evaluate the per-primitive variances $v_k=\vpsi_k^\top\mSigma_k\vpsi_k$ at the query's
direction and time and composite them with the squared weights $w_k(q)^2$, as for per-sample variances
in~\cite{pan2022activenerf}:
\begin{equation}
  v(q)=\textstyle\sum_k w_k(q)^2\,v_k.
\label{eq:blocks}
\end{equation}
Keeping one block per primitive and treating correlated pixels as independent leave the absolute scale of $h$
open; calibration sets it (\cref{sec:calib}).
Both $\mathbf{H}_k$ and $v(q)$ are estimated with the backbone's own rasterizer, by gathering and splatting randomly
signed probes~\cite{martens2012training,alon1996space} (details in the supplement), and $v(q)$ is smoothed over covered pixels like the targets.

\textbf{Predictive variance.}
As in a prediction interval, the variance of a new residual at a query pixel $q$ adds the posterior variance $v(q)$ of the correction to the noise:
\begin{equation}
  e^{\eta(q)}+v(q)=e^{\eta(q)}(1+h(q)),\quad h(q)=v(q)\,e^{-\eta(q)}.
  \label{eq:predictive}
\end{equation}
Where the training views observe the primitives well, $h(q)$ is small and the estimate stays at the error level;
where they constrain the correction poorly, $h(q)$ grows, where the error level alone tends to underestimate the
error.

\input{tables/dense_mono}

\begin{figure*}[t]
  \centering
  \includegraphics[width=\textwidth]{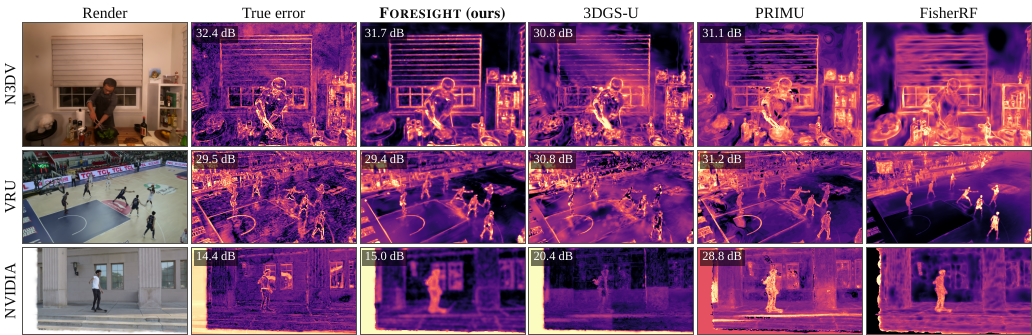}
  \vspace{-1.5em}
  \caption{Uncertainty estimated by each method and true error of the rendering on dense rigs (N3DV, VRU) and
  monocular capture (NVIDIA). Each map is on its own log scale (brighter is higher); labels give the true PSNR and
  each method's estimate (none for FisherRF). On the dense rigs, \ours follows the structure of the error and
  estimates the PSNR to within 0.7\,dB. From a single camera, the baselines estimate the PSNR 6 to 14\,dB too high. More views in the supplement.}
  \label{fig:qual}
\end{figure*}

\subsection{What the linear color model leaves open}
\label{sec:open}

\textbf{Beyond the observed time span.}\label{sec:time}
The error level and the leverage are learned within the observed span; past it, each primitive keeps its
end-of-span error level and leverage, which change with the viewpoint and the backbone's motion, but not with the horizon $\delta=t-1$.
A forecast frame is wrong mostly where the scene moves, and no ground truth exists there.
We therefore estimate the error of the simplest forecast, holding the last frame, from the motion within the span:
for each training camera, the squared change between the backbone's renders at times $s$ and $s+l$, over
all such pairs, at lags $l$ up to a quarter of the span, so that three quarters of it still give pairs.
We gather this change onto the primitives as $\gamma_k(l)$, the weighted mean change
of the pixels each covers.
We interpolate $\gamma_k$ between the lags to $\delta$, extend it linearly beyond them, and splat it with
the weights at the end of the span, $w_k(p;1)$: $\sigma^2_{\mathrm{time}}(p,\delta)=\sum_kw_k(p;1)\,\gamma_k(\delta)$,
bounded by the ceiling (\cref{sec:ceiling}).
Like dynamical-system forecasters~\cite{asiimwe2025evogs}, this assumes that the scene keeps changing as it did
within the span, here also in the same places.
The linear extension can make the estimate pessimistic at long horizons, which we discuss in
\cref{sec:limitations}.
The time term is zero within the span and has no fitted scale.

\textbf{Know-nothing ceiling.}\label{sec:ceiling}
No estimate should exceed the error of knowing nothing.
If the true color were an uninformed draw from the scene's colors, with mean $\bar{\mathbf c}$ and variance $\vnu$
over the training images, the render's expected squared error would be
$\sigma^2_{\mathrm{ceil}}(p)=\frac13\bigl(\lVert\hat I(p)-\bar{\mathbf c}\rVert^2+\mathbf 1^\top\vnu\bigr)$, which
caps the estimate.
Where little geometry was reconstructed, $A(p)<\frac12$, the leverage sees few primitives and stays small
although nothing there was observed; there, the estimate moves toward the ceiling (\cref{eq:final}).

\textbf{Calibration and full estimate.}
\label{sec:calib}
We fit four scalars by the Gaussian negative log-likelihood of the residuals of one held-out camera: first $a$, $d$ and $\kappa$ on the uncapped estimate, then
$\lambda_{\mathrm{ceil}}$:
\begin{equation}\redmath
\begin{split}
  \sigma^2_{\mathrm{cap}} &= \min\Bigl(
   \underbrace{e^{a\,\eta+d}}_{\text{error level}}
   \underbrace{(1+\kappa\,h)}_{\text{leverage}}
   + \underbrace{\sigma^2_{\mathrm{time}}}_{\text{time}},\
   \underbrace{\sigma^2_{\mathrm{ceil}}}_{\text{ceiling}}\Bigr), \\
  \sigma^2 &= (1-\lambda_{\mathrm{ceil}}(p))\,\sigma^2_{\mathrm{cap}} + \lambda_{\mathrm{ceil}}(p)\,\sigma^2_{\mathrm{ceil}}, 
\end{split}
\label{eq:final}
\end{equation}
with $\kappa\ge0$ and $\lambda_{\mathrm{ceil}}(p)=\mathbb{1}_{A(p) < \frac12}\cdot\hat{\lambda}_{\mathrm{ceil}}$, where $\hat{\lambda}_{\mathrm{ceil}}\in[0,1]$.
With $a=\kappa=1$, $d=0$, no time term and no ceiling ($\hat\lambda_{\mathrm{ceil}}=0$, no minimum), \cref{eq:final} reduces to the prediction interval
$e^{\eta}(1+h)$ of \cref{eq:predictive}.
The held-out camera sets only these scalars; every term they scale is computed without it.

%% file: figs/method/method.tex
\begingroup
\definecolor{mfEmp}{RGB}{22,150,68}    %
\definecolor{mfEpi}{RGB}{0,150,196}    %
\definecolor{mfTime}{RGB}{190,60,90}   %
\definecolor{mfCeil}{RGB}{122,88,160}  %
\definecolor{mfCal}{RGB}{212,112,24}   %
\definecolor{mfBb}{RGB}{84,104,128}    %
\def\mftw{1.5}\def\mfth{1.125}\def\mflh{1.08}\def\mfmw{1.32}\def\mfmh{0.99}
\def\mfya{0}\def\mfyb{-1.2}\def\mfyd{-2.4}\def\mfyc{-3.6}\def\mfyt{0.6}\def\mfyl{-4.2}
\def\mfwf{4.15}\def\mfwp{2.05}
\def\mfxf{2.4}\def\mfwc{3.45}\def\mfwo{2.25}\def\mfoh{1.6875}\def\mfyf{-1.15}\def\mfycal{-3.05}
\pgfmathsetmacro{\mfxfe}{\mfxf+\mfwf}
\pgfmathsetmacro{\mfxp}{\mfxfe+0.3}
\pgfmathsetmacro{\mfxpe}{\mfxp+\mfwp}
\pgfmathsetmacro{\mfxpc}{\mfxp+\mfwp/2}
\pgfmathsetmacro{\mfxm}{\mfxpe+0.3}
\pgfmathsetmacro{\mfme}{\mfxm+\mfmw}
\pgfmathsetmacro{\mfxc}{\mfme+0.83}
\pgfmathsetmacro{\mfxcc}{\mfxc+\mfwc/2}
\pgfmathsetmacro{\mfxo}{\mfxc+\mfwc+0.3}
\begin{tikzpicture}[x=1cm, y=1cm, line cap=round, line join=round,
  font=\fontsize{7}{8}\selectfont,
  >={Stealth[length=1.5mm, width=1.2mm]},
  box/.style={rounded corners=1.5pt, line width=0.5pt, draw=#1, fill=#1!9},
  flow/.style={->, line width=0.6pt, rounded corners=1.5pt, draw=#1},
  flow/.default={black!60},
  thumb/.style={inner sep=0pt, outer sep=0pt, line width=0.7pt, draw=#1, fill=black!6},
  thumb/.default={black!50},
  tag/.style={anchor=south west, fill=white, fill opacity=0.85, text opacity=1, inner sep=1.3pt, outer sep=0.6pt},
  head/.style={anchor=south, font=\fontsize{7}{8}\selectfont\bfseries, text=black!55, inner sep=2pt, text height=1.5ex, text depth=0.5ex},
  txt/.style={anchor=west, align=left, inner sep=0pt},
  tiny/.style={font=\fontsize{6}{7}\selectfont, inner sep=0pt},
  step/.style={circle, draw=#1, fill=white, line width=0.5pt, inner sep=0pt, minimum size=0.27cm,
               font=\fontsize{6}{6}\selectfont\bfseries, text=#1},
  sec/.style={anchor=north east, font=\fontsize{6}{7}\selectfont, text=black!45, inner sep=0pt},
  cam/.pic={\filldraw[pic actions] (0,0) -- (0.17,0.085) -- (0.17,-0.085) -- cycle;
            \filldraw[pic actions] (-0.1,-0.055) rectangle (0,0.055);},
  flake/.pic={\foreach \ang in {0,60,...,300}
      \draw[pic actions] (0,0) -- (\ang:0.13) (\ang:0.075) -- ++(\ang+50:0.05) (\ang:0.075) -- ++(\ang-50:0.05);}]

  \path[box=mfBb] (0,\mfyt) rectangle (1.9,-1.2);
  \begin{scope}[shift={(0.88,0.02)}]
    \foreach \gx/\gy/\gr/\ga/\gb in {-0.32/0.12/25/0.2/0.1, 0.02/0.22/-35/0.17/0.085, 0.3/0.06/60/0.16/0.08,
                                     -0.12/-0.14/-10/0.2/0.095, 0.2/-0.2/35/0.15/0.08}
      \filldraw[draw=mfBb, fill=mfBb!35, line width=0.4pt, rotate around={\gr:(\gx,\gy)}]
        (\gx,\gy) ellipse [x radius=\ga, y radius=\gb];
  \end{scope}
  \pic[draw=mfBb, line width=0.5pt] at (1.63,0.33) {flake};
  \node[align=center, inner sep=0pt] at (0.95,-0.72) {{\bfseries\color{mfBb}Frozen backbone}\\dynamic Gaussian\\splatting};
  \node[thumb, anchor=south west] (train3) at ({0.06+0.26},{-3.95+0.26}) {\includegraphics[width=\mftw cm,height=\mfth cm]{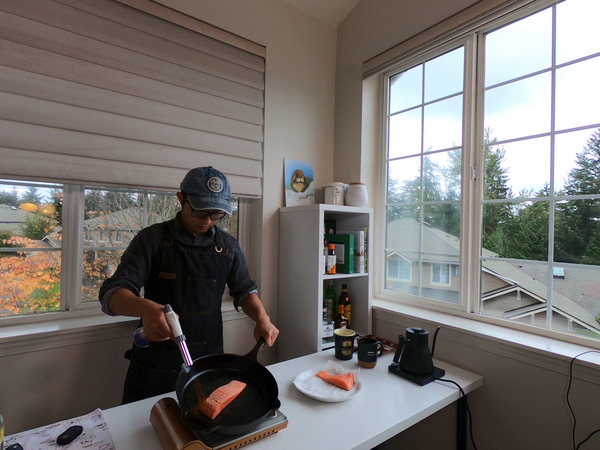}};
  \node[thumb, anchor=south west] (train2) at ({0.06+0.13},{-3.95+0.13}) {\includegraphics[width=\mftw cm,height=\mfth cm]{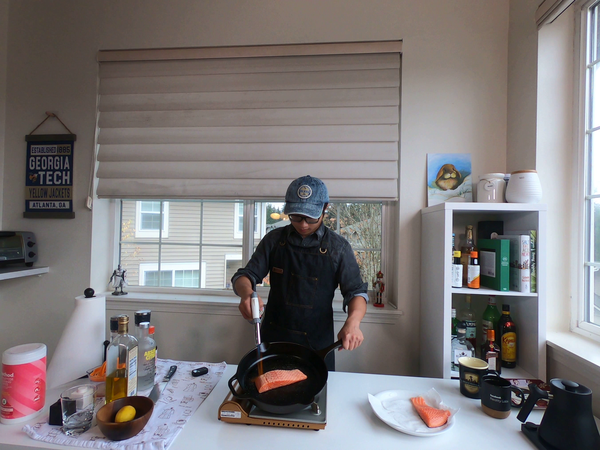}};
  \node[thumb, anchor=south west] (train) at ({0.06+0},{-3.95+0}) {\includegraphics[width=\mftw cm,height=\mfth cm]{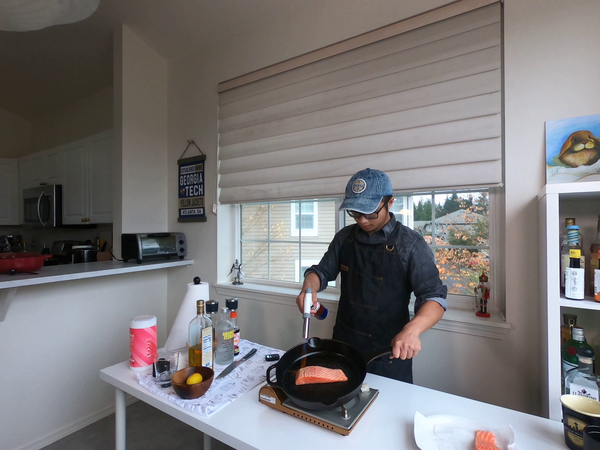}};
  \node[inner sep=0pt] at (0.95,{\mfyl+0.08}) {training images $I_f$};
  \draw[line width=0.6pt, draw=black!60, rounded corners=1.5pt] (1.9,-0.3) -- (2.15,-0.3)
    (1.85,-3.05) -- (2.15,-3.05) -- (2.15,\mfya);
  \foreach \y in {\mfya,\mfyb} \draw[flow] (2.15,\y) -- (\mfxf,\y);
  \draw[draw=white, line width=2.6pt] (2.03,\mfyd) -- (2.27,\mfyd);   %
  \draw[flow] (1.7,-1.2) -- (1.7,\mfyd) -- (\mfxf,\mfyd);
  \draw[flow] (1.85,\mfyc) -- (\mfxf,\mfyc);

  \foreach \y/\c in {\mfya/mfEmp, \mfyb/mfEpi, \mfyd/mfTime, \mfyc/mfCeil}
    \path[box=\c] (\mfxf,{\y+\mflh/2}) rectangle (\mfxfe,{\y-\mflh/2});
  \def\mftxtemp{log error level $\eta_k(\boldsymbol\omega,t)$ per primitive,\\fitted to the squared training residuals}
  \def\mftxtepi{how well the training views constrain\\a color correction of each primitive}
  \def\mftxttime{scene change $\gamma_k$ in the backbone's own\\renders, gathered per primitive}
  \def\mftxtceil{error of an uninformed guess, from the\\training color statistics $\bar{\mathbf c},\boldsymbol\nu$}
  \foreach \y/\c/\n/\t/\s/\b in {\mfya/mfEmp/1/Error level/3.2/\mftxtemp, \mfyb/mfEpi/2/Leverage/3.3/\mftxtepi,
                                 \mfyd/mfTime/3/Time term/3.4/\mftxttime, \mfyc/mfCeil/4/Ceiling/3.4/\mftxtceil} {
    \node[step=\c] at ({\mfxf+0.24},{\y+0.27}) {\n};
    \node[txt] at ({\mfxf+0.45},{\y+0.27}) {\bfseries\color{\c}\t};
    \node[txt] at ({\mfxf+0.15},{\y-0.13}) {\b};
    \node[sec] at ({\mfxfe-0.08},{\y+\mflh/2-0.07}) {\S\,\s};
  }

  \draw[flow=mfEmp] ({\mfxfe-0.6},{\mfya-\mflh/2+0.12}) -- ({\mfxfe-0.6},{\mfyb+\mflh/2-0.1});

  \path[box=mfBb] (\mfxp,\mfyt) rectangle (\mfxpe,\mfyl);
  \pic[draw=mfBb, line width=0.5pt] at ({\mfxp+0.22},0.38) {flake};
  \begin{scope}[shift={(\mfxpc,-0.75)}]
    \foreach \gx/\gy/\gr/\ga/\gb/\gc in {0/0.05/20/0.22/0.11/mfBb, 0.12/0.3/-30/0.17/0.085/mfBb,
          -0.24/0.24/60/0.16/0.085/mfBb, 0.14/-0.2/10/0.2/0.1/mfBb, -0.26/-0.12/-40/0.17/0.09/mfBb,
          0.42/0.04/70/0.14/0.075/mfEpi, 0.4/0.34/15/0.12/0.065/mfEpi}
      \filldraw[draw=\gc, fill=\gc!40, line width=0.4pt, rotate around={\gr:(\gx,\gy)}]
        (\gx,\gy) ellipse [x radius=\ga, y radius=\gb];
    \foreach \t in {185,210,235,260,285}
      \pic[draw=black!50, fill=black!12, line width=0.35pt, rotate=\t+180] at (\t:0.85) {cam};
    \fill[black, opacity=0.07] (30:0.9) -- ++(190:0.8) -- ($(30:0.9)+(230:0.8)$) -- cycle;
    \draw[densely dashed, line width=0.35pt, black!70] (30:0.9) -- ++(190:0.8) (30:0.9) -- ++(230:0.8);
    \pic[draw=black, fill=black!75, line width=0.35pt, rotate=210] at (30:0.95) {cam};
    \node[tiny, anchor=south] at (0.66,0.68) {query};
    \node[tiny, text=mfEpi] at (0.72,-0.36) {high $h$};
    \node[tiny] at (-0.12,-1.1) {training views};
  \end{scope}
  \node[thumb] (render) at (\mfxpc,{\mfyl+0.1+\mfth/2}) {\includegraphics[width=\mftw cm,height=\mfth cm]{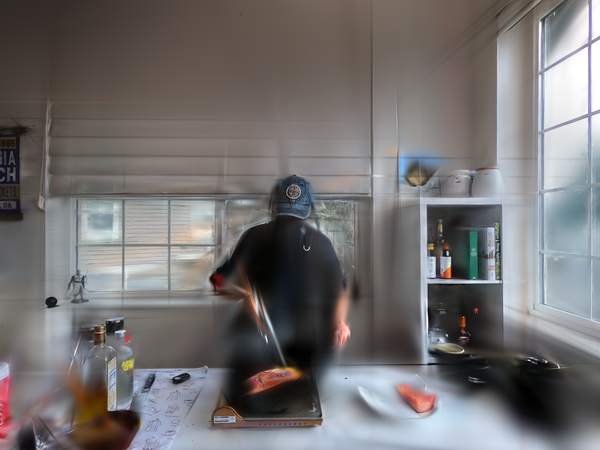}};
  \node[tag] at (render.south west) {render $\hat I$};

  \node[thumb=mfEmp] (eta) at ({\mfxm+\mfmw/2},\mfya) {\includegraphics[width=\mfmw cm,height=\mfmh cm]{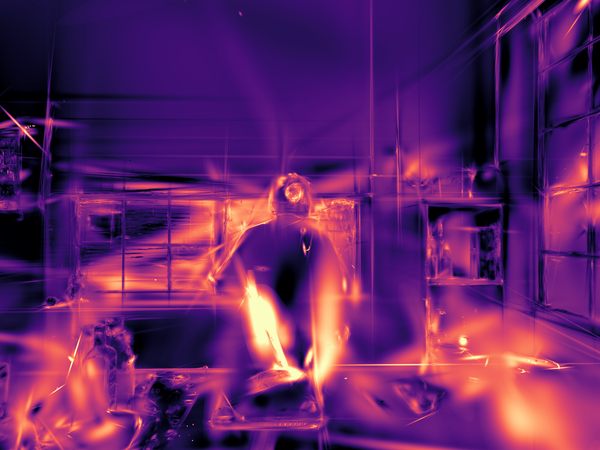}};
  \node[thumb=mfEpi] (lev) at ({\mfxm+\mfmw/2},\mfyb) {\includegraphics[width=\mfmw cm,height=\mfmh cm]{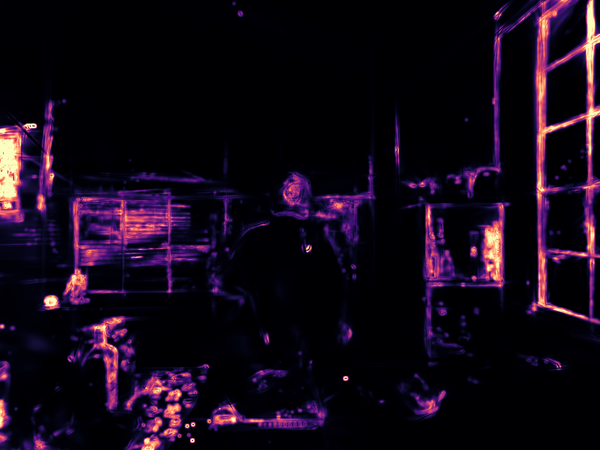}};
  \node[thumb=mfTime] (time) at ({\mfxm+\mfmw/2},\mfyd) {\includegraphics[width=\mfmw cm,height=\mfmh cm]{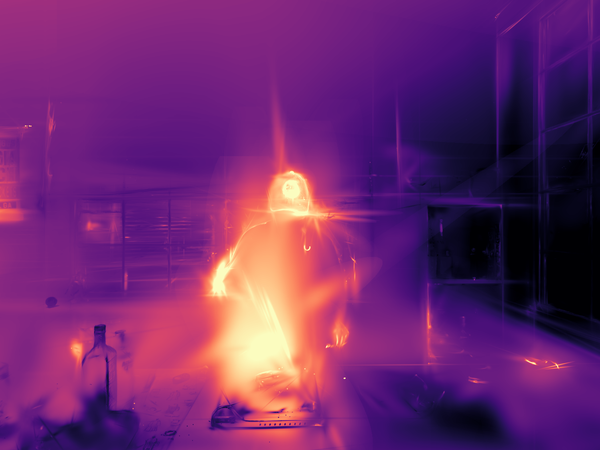}};
  \node[thumb=mfCeil] (ceil) at ({\mfxm+\mfmw/2},\mfyc) {\includegraphics[width=\mfmw cm,height=\mfmh cm]{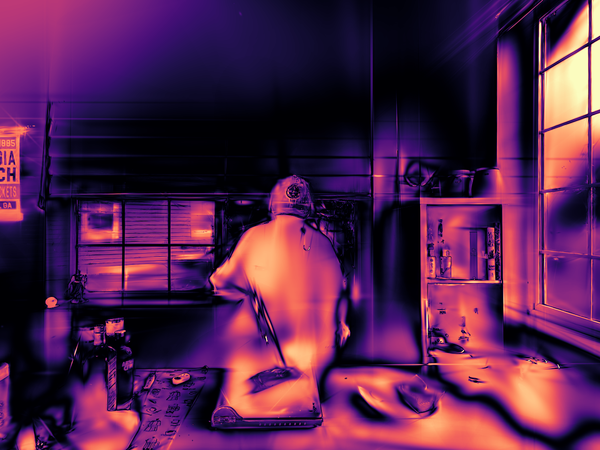}};
  \foreach \y/\c/\l in {\mfya/mfEmp/{$e^{a\eta+d}$}, \mfyb/mfEpi/{$1+\kappa h$},
                        \mfyd/mfTime/{$\sigma^2_{\mathrm{time}}$}, \mfyc/mfCeil/{$\sigma^2_{\mathrm{ceil}}$}} {
    \node[rotate=90, inner sep=0pt, font=\fontsize{6}{7}\selectfont, text=\c] at ({\mfme+0.19},\y) {\l};   %
  }
  \foreach \y/\c in {\mfya/mfEmp, \mfyb/mfEpi, \mfyd/mfTime, \mfyc/mfCeil} {
    \draw[flow=\c] (\mfxfe,\y) -- (\mfxp,\y);
    \draw[flow=\c] (\mfxpe,\y) -- (\mfxm,\y);
  }

  \def\mfcalpars{a,d,\kappa,\lambda}
  \node[box=black, draw=black!60, fill=white, minimum width=\mfwc cm, minimum height=1.2cm, inner sep=0pt,
        align=center, font=\fontsize{8}{11}\selectfont] (comb) at (\mfxcc,\mfyf)
    {{\color{mfEmp}error level} $\times$ (1 + {\color{mfEpi}leverage})\\+ {\color{mfTime}time term} past the span,\\capped at the {\color{mfCeil}ceiling}};
  \draw[flow=mfEmp] ({\mfme+0.33},\mfya) -- ({\mfme+0.56},\mfya) -- ({\mfme+0.56},{\mfyf+0.33}) -- (\mfxc,{\mfyf+0.33});
  \draw[flow=mfEpi] ({\mfme+0.33},\mfyb) -- ({\mfme+0.44},\mfyb) -- ({\mfme+0.44},{\mfyf+0.11}) -- (\mfxc,{\mfyf+0.11});
  \draw[flow=mfTime] ({\mfme+0.33},\mfyd) -- ({\mfme+0.44},\mfyd) -- ({\mfme+0.44},{\mfyf-0.11}) -- (\mfxc,{\mfyf-0.11});
  \draw[flow=mfCeil] ({\mfme+0.33},\mfyc) -- ({\mfme+0.56},\mfyc) -- ({\mfme+0.56},{\mfyf-0.33}) -- (\mfxc,{\mfyf-0.33});
  \node[box=mfCal, minimum width=\mfwc cm, minimum height=1cm, inner sep=0pt] (cal) at (\mfxcc,\mfycal) {};
  \pic[draw=mfCal, fill=mfCal!35, line width=0.4pt, scale=1.3] at ({\mfxc+0.27},\mfycal) {cam};
  \node[step=mfCal] at ({\mfxc+0.79},{\mfycal+0.28}) {5};
  \node[txt] at ({\mfxc+1.0},{\mfycal+0.28}) {\bfseries\color{mfCal}Calibration};
  \node[txt] at ({\mfxc+0.65},{\mfycal-0.15}) {${\color{mfCal}\mfcalpars}$ fitted on a\\held-out camera};
  \node[sec] at ({\mfxc+\mfwc-0.08},{\mfycal+0.43}) {\S\,3.4};
  \draw[flow=mfCal] (cal.north) -- (comb.south);

  \node[thumb=black!75, anchor=west] (sigma) at (\mfxo,\mfyf) {\includegraphics[width=\mfwo cm,height=\mfoh cm]{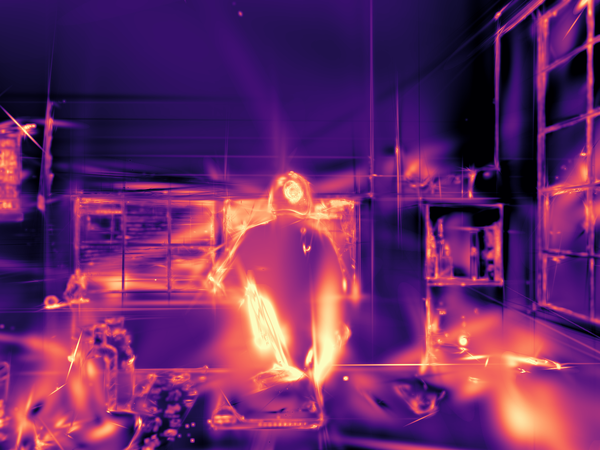}};
  \draw[flow] (comb.east) -- (sigma.west);
  \node[anchor=north, align=center, inner sep=0pt, outer sep=2.5pt] at (sigma.south)
    {estimated $\sigma^2(p)$\\[1pt]{\fontsize{6.5}{7.5}\selectfont est.\ PSNR 20.4\,dB}\\{\fontsize{6.5}{7.5}\selectfont (true 20.5\,dB)}};

  \node[head] at (0.95,\mfyt) {Inputs};
  \node[head] at ({\mfxf+\mfwf/2},\mfyt) {Fit once per scene};
  \node[head] at (\mfxpc,\mfyt) {Query viewpoint, time};
  \node[head] at ({\mfxm+\mfmw/2+0.16},\mfyt) {Per-pixel maps};   %
  \node[head] at (\mfxcc,\mfyt) {Calibrate};
  \node[head] at ({\mfxo+\mfwo/2},\mfyt) {Estimate};
\end{tikzpicture}%
\endgroup

%% file: tables/dense_mono.tex
\begin{table*}[t]
  \centering
  \small
  \setlength{\tabcolsep}{5pt}
  \caption{Comparison of photometric uncertainty on dense multi-view and monocular capture.
  \ours achieves the \firstbox{best} error localization (AUSE), calibration (NLL) and PSNR estimate ($\Delta$PSNR) in
  all settings; \secondbox{second best} is also highlighted.
  $\Delta$PSNR is not reported for FisherRF, as its score is parameter uncertainty rather than an estimate of the
  photometric error.}
  \label{tab:main}
  \vspace{-0.5em}
  \begin{tabular}{@{}l*{9}{c}}  %
    \toprule
    & \multicolumn{3}{c}{N3DV~\cite{li2022neural} (dense)}
    & \multicolumn{3}{c}{VRU~\cite{wu2025swift4d} (dense, large motion)}
    & \multicolumn{3}{c}{NVIDIA~\cite{yoon2020novel} (monocular)} \\
    Backbone & \multicolumn{3}{c}{4DGaussians~\cite{wu2024_4dgaussians}}
    & \multicolumn{3}{c}{ATGS~\cite{wu2026atgs}}
    & \multicolumn{3}{c}{Shape of Motion~\cite{wang2025shape}} \\
    \cmidrule(lr){2-4} \cmidrule(lr){5-7} \cmidrule(l){8-10}
    Method & AUSE$\downarrow$ & NLL$\downarrow$ & $\Delta$PSNR$\downarrow$ & AUSE$\downarrow$ & NLL$\downarrow$ & $\Delta$PSNR$\downarrow$ & AUSE$\downarrow$ & NLL$\downarrow$ & $\Delta$PSNR$\downarrow$ \\
    \midrule
    FisherRF~\cite{jiang2024fisherrf} & 0.581 & $-$2.192 & -- & 0.576 & $-$1.626 & -- & 1.504 & 4.794 & -- \\
    3DGS-U~\cite{galappaththige2026predictive} & \second{0.251} & \second{$-$2.352} & 1.11 & \second{0.316} & \second{$-$1.835} & \second{1.23} & \second{0.161} & \second{$-$1.065} & \second{4.79} \\
    PRIMU~\cite{gottwald2026primu} & 0.329 & $-$2.269 & \second{0.76} & 0.338 & $-$1.647 & 2.26 & 0.626 & 1.669 & 6.76 \\
    \arrayrulecolor{black!25}\midrule\arrayrulecolor{black}
    \ours (ours) & \first{0.224} & \first{$-$2.414} & \first{0.66} & \first{0.231} & \first{$-$1.894} & \first{1.12} & \first{0.112} & \first{$-$1.225} & \first{0.61} \\
    \bottomrule
  \end{tabular}
\end{table*}

%% file: sec/04_experiments.tex
\section{Experiments}
\label{sec:experiments}

\noindent\textbf{Datasets and evaluation settings.}
We evaluate \ours on five backbones under five settings; \cref{sec:ablation} shows which term of \ours carries the gain in each.
For a dense multi-view rig, we evaluate on N3DV~\cite{li2022neural} following the protocol of
4DGaussians~\cite{wu2024_4dgaussians}.
For a dense rig with large motion, we evaluate on VRU Basketball~\cite{wu2025swift4d} following ATGS~\cite{wu2026atgs}.
For a sparse rig, we follow the four-camera setting of 4C4D~\cite{zhou20264c4d} on N3DV.
For monocular capture, we evaluate on NVIDIA Dynamic Scenes~\cite{yoon2020novel} following Shape of
Motion~\cite{wang2025shape}.
Finally, for future frames beyond the observed time span, we use the default of the official EvoGS~\cite{asiimwe2025evogs} implementation on N3DV, which observes the first half of each sequence and extrapolates the second.
In each setting, the method whose protocol we follow serves as the backbone, and one camera, used for neither
training nor testing, is held out for calibration (\cref{sec:calib}); for future frames, only its first half is used.

\noindent\textbf{Baselines.}
Uncertainty-aware dynamic methods~\cite{kim20244d,kim2026gp,guo2026uncertainty} estimate the uncertainty of the
model trained with them, not of an already trained one, so we adopt three post-hoc methods for static 3DGS with
public code and adapt them to the backbone's primitives deformed to each frame's time.
\mbox{3DGS-U}~\cite{galappaththige2026predictive} and PRIMU~\cite{gottwald2026primu} estimate the error of a
rendering, while FisherRF~\cite{jiang2024fisherrf} quantifies the uncertainty of the parameters; details of the
adaptations are in the supplement.

\noindent\textbf{Metrics.}
Following prior work~\cite{galappaththige2026predictive,gottwald2026primu,kim2026gp}, we measure how well the
estimated uncertainty localizes the per-pixel squared error by the area under the sparsification error
(AUSE)~\cite{ilg2018uncertainty}, computed per view and averaged.
As \ours estimates the variance of the rendering error itself, we also evaluate its magnitude by the Gaussian negative
log-likelihood (NLL) of the residuals.
For the PSNR estimated from the mean of $\sigma^2$ over a view,
$\widehat{\mathrm{PSNR}}=-10\log_{10}\overline{\sigma^2}$ (\cref{sec:method}), we report its mean absolute error
to the true PSNR ($\Delta$PSNR, in dB).
For a fair comparison, all methods are applied to the same frozen backbone, use only its training views, learn
from the squared rendering error of these views (except FisherRF, which measures parameter uncertainty), and
map their scores to variances by the same log-affine calibration on the held-out camera (further metrics in the
supplement).

\subsection{Dense and monocular capture}
\label{sec:results_main}

\Cref{tab:main,fig:qual} compare \ours with the baselines on the dense rigs and on monocular capture.
\ours is best on all three metrics in all three settings. On the dense rigs, every test camera is surrounded by training cameras, and the photometric baselines come closest.
\ours still lowers the AUSE of the best baseline by 11\% on N3DV and by 27\% on VRU, whose players move fast and far, and estimates the PSNR with mean absolute errors of 0.66\,dB and 1.12\,dB, respectively.

On monocular capture, \ours lowers the AUSE of the best baseline, \mbox{3DGS-U}, by 30\% and estimates the PSNR
to within 0.61\,dB, whereas the baselines miss it by 4.8--6.8\,dB.
The gap arises in the parts of each test view that the single training camera never observed; on the pixels it
also sees, \mbox{3DGS-U}, PRIMU and \ours all estimate the PSNR to within 1.1--1.3\,dB (supplement).
In the unobserved parts, PRIMU and FisherRF remain as confident as on the training views and underestimate the
squared error by a factor of two to three even after calibration, whereas \mbox{3DGS-U}, with its option that makes
pixels without geometry maximally uncertain, overestimates it by a factor of 2.5.
\ours assigns these parts its ceiling (\cref{sec:ceiling}), and its leverage lowers AUSE by a further 15\%
(supplement).

\input{tables/sparse}
\input{tables/time}
\input{tables/ablation}

\subsection{Sparse rig and out-of-distribution views}
\label{sec:results_sparse}

\begin{figure}[t]
  \centering
  \includegraphics[width=\columnwidth]{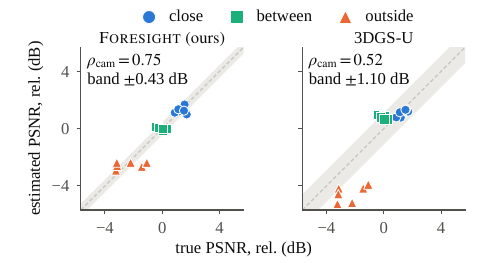}
  \vspace{-1.6em}
  \caption{Estimating the quality of novel views on the sparse rig. Each point is one camera group of one scene:
  the mean true and estimated PSNR of its test cameras, relative to the scene mean. $\rho_{\mathrm{cam}}$: Spearman
  correlation of estimated and true PSNR over a scene's test cameras, each averaged over its frames;
  mean over scenes. Band: mean absolute deviation from the diagonal.}
  \label{fig:view_ranking}
\end{figure}

Sparse rigs make dynamic capture affordable, but leave most novel views far from the training data.
We group the test cameras into those close to a training camera, those between the training cameras, and those
outside their span (definitions in the supplement).
\ours achieves the best AUSE and NLL (\cref{tab:sparse}), with 19\% lower AUSE than the best baseline, and remains
best on both in each group (supplement).
Its PSNR estimate is second to PRIMU's (1.73 \vs 1.67\,dB).

\noindent\textbf{Ranking novel views.}
On a sparse rig, some viewpoints render several
decibels worse than others, and without ground truth at a novel viewpoint, one cannot tell which.
Across the test views of each scene, the Spearman correlation between estimated and true PSNR is 0.69 for \ours,
against 0.52 for \mbox{3DGS-U} and 0.43 for the acquisition score of FisherRF.
The estimates of \ours lie close to the diagonal for all three groups of test cameras (\cref{fig:view_ranking}), deviating by 0.43\,dB on average, against 1.10\,dB
for \mbox{3DGS-U}, which hardly separates close from between views and estimates those outside the span far too low.

\subsection{Future frames and out-of-distribution time}
\label{sec:results_time}

\begin{figure}[t]
  \centering
  \includegraphics[width=\columnwidth]{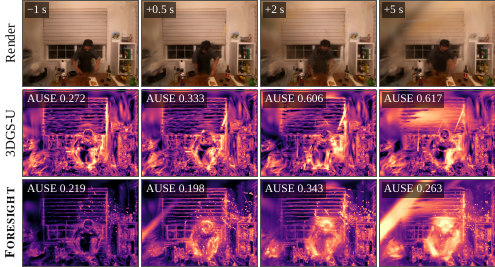}
  \caption{Future frames (N3DV, EvoGS) at the horizons of \cref{fig:teaser}, with the uncertainty of \mbox{3DGS-U},
  the best baseline in \cref{tab:time}, and of \ours. Each row of maps is on one log scale; labels give the view's
  AUSE.}
  \label{fig:future}
\end{figure}

Dynamic models are increasingly used to forecast upcoming
frames~\cite{asiimwe2025evogs}, where no ground truth exists yet and the error grows with the horizon.
On the future half of each sequence, \ours is best on all three metrics (\cref{tab:time}).
Its AUSE on future frames matches that on observed ones (0.234 \vs 0.231), whereas that of the best baseline,
\mbox{3DGS-U}, grows from 0.305 to 0.360.
\Cref{fig:future} shows this on one scene: over five seconds, the AUSE of \mbox{3DGS-U} more than doubles, from
0.272 to 0.617, whereas that of \ours stays below 0.35, as its time term marks the moving person.

\subsection{Component analysis}
\label{sec:ablation}

\Cref{tab:ablation} adds the components of \ours in turn, with the same hyperparameters in every setting.
Calibration corrects the level and spread of the error level and improves NLL; being monotone, it leaves AUSE
unchanged.
Leverage and ceiling enter together, as the ceiling bounds the estimate and takes over where little geometry
was reconstructed (\cref{sec:ceiling}).
Together they improve AUSE and NLL in every setting within the observed time span, least where every test
camera is surrounded by training cameras (AUSE lower by 3\% and 11\% on the dense rigs) and most where the test
views lie far from them (21\% on the sparse rig, 37\% on monocular capture); on the sparse rig, they lower
AUSE by 33\% at the viewpoints outside its span, against 18\% at those close to the training cameras, in all six
scenes.
On the dense rigs, the error level alone already has a lower AUSE than the best baseline (0.231 \vs 0.251 on N3DV,
0.260 \vs 0.316 on VRU).
Zero within the observed span, the time term lowers AUSE on future frames by 14\% and NLL from $-1.251$ to
$-1.359$.

\input{tables/swap}
\noindent\textbf{Pairing an error regression with parameter uncertainty.}
\Cref{tab:swap} multiplies the error level of \mbox{3DGS-U} and the error level of \ours by one plus a
leverage on the sparse rig.
A leverage built from FisherRF's per-primitive information leaves both error levels unchanged (AUSE 0.357 \vs 0.358
for \mbox{3DGS-U}).
Our leverage lowers the AUSE of \mbox{3DGS-U} to 0.321 and that of the error level of \ours to 0.292.
Naively pairing an error regression with parameter uncertainty thus does not improve the estimate; our formulation of
the leverage (\cref{sec:leverage}) does.

%% file: tables/sparse.tex
\begin{table}[t]
  \centering
  \small
  \setlength{\tabcolsep}{4pt}
  \caption{Sparse rig: N3DV with four training cameras, backbone
  4C4D~\cite{zhou20264c4d}.
  $\rho_{\mathrm{view}}$: Spearman correlation of estimated and true PSNR across a scene's test views (every
  camera at every frame); FisherRF ranks them by its acquisition score.
  \ours achieves the \firstbox{best} AUSE, NLL and $\rho_{\mathrm{view}}$, and \secondbox{second-best}
  $\Delta$PSNR.}
  \label{tab:sparse}
  \vspace{-0.5em}
  \begin{tabular}{@{}l*{4}{c}}  %
    \toprule
    Method & AUSE$\downarrow$ & NLL$\downarrow$ & $\Delta$PSNR$\downarrow$ & $\rho_{\mathrm{view}}{\uparrow}$ \\
    \midrule
    FisherRF~\cite{jiang2024fisherrf} & 0.491 & $-$1.184 & -- & 0.43 \\
    3DGS-U~\cite{galappaththige2026predictive} & \second{0.358} & \second{$-$1.237} & 1.87 & \second{0.52} \\
    PRIMU~\cite{gottwald2026primu} & 0.407 & $-$1.175 & \first{1.67} & 0.37 \\
    \arrayrulecolor{black!25}\midrule\arrayrulecolor{black}
    \ours (ours) & \first{0.290} & \first{$-$1.270} & \second{1.73} & \first{0.69} \\
    \bottomrule
  \end{tabular}
\end{table}

%% file: tables/time.tex
\begin{table}[t]
  \centering
  \small
  \setlength{\tabcolsep}{4pt}
  \caption{Comparison of photometric uncertainty on future frames: N3DV, backbone EvoGS~\cite{asiimwe2025evogs}
  trained on the first half of each sequence and evaluated on the second.
  \ours achieves the \firstbox{best} AUSE, NLL and $\Delta$PSNR; \secondbox{second best} is also highlighted.}
  \label{tab:time}
  \vspace{-0.5em}
  \begin{tabular}{@{}l*{3}{c}}  %
    \toprule
    Method & AUSE$\downarrow$ & NLL$\downarrow$ & $\Delta$PSNR$\downarrow$ \\
    \midrule
    FisherRF~\cite{jiang2024fisherrf} & 0.567 & $-$0.696 & -- \\
    3DGS-U~\cite{galappaththige2026predictive} & \second{0.360} & \second{$-$1.029} & \second{2.48} \\
    PRIMU~\cite{gottwald2026primu} & 0.419 & $-$0.784 & 3.63 \\
    \arrayrulecolor{black!25}\midrule\arrayrulecolor{black}
    \ours (ours) & \first{0.234} & \first{$-$1.359} & \first{2.03} \\
    \bottomrule
  \end{tabular}
\end{table}

%% file: tables/ablation.tex
\begin{table*}[t]
  \centering
  \small
  \setlength{\tabcolsep}{5pt}
  \caption{Component analysis of \ours on the settings of \cref{tab:main,tab:sparse,tab:time}, adding the
  components in turn; the last row is \ours.
  Gray: identical to the row above by construction, as calibration is monotone and the time term is zero within the
  observed time span. On future frames, only the time term is ablated.
  Each component improves NLL wherever it acts, and all but calibration also improve AUSE.}
  \label{tab:ablation}
  \begin{tabular}{@{}l*{10}{c}@{}}
    \toprule
    & \multicolumn{8}{c}{Within the observed time span} & \multicolumn{2}{c}{Beyond it} \\
    \cmidrule(lr){2-9} \cmidrule(l){10-11}
    & \multicolumn{2}{c}{N3DV (dense)} & \multicolumn{2}{c}{VRU (dense)} & \multicolumn{2}{c}{N3DV (sparse)} & \multicolumn{2}{c}{NVIDIA (monocular)} & \multicolumn{2}{c}{N3DV (future)} \\
    \cmidrule(lr){2-3} \cmidrule(lr){4-5} \cmidrule(lr){6-7} \cmidrule(lr){8-9} \cmidrule(l){10-11}
    Component & AUSE$\downarrow$ & NLL$\downarrow$ & AUSE$\downarrow$ & NLL$\downarrow$ & AUSE$\downarrow$ & NLL$\downarrow$ & AUSE$\downarrow$ & NLL$\downarrow$ & AUSE$\downarrow$ & NLL$\downarrow$ \\
    \midrule
    Error level & 0.231 & $-$1.468 & 0.260 & $-$1.218 & 0.368 & 17.462 & 0.178 & 31.041 &  &  \\
    + calibration ($a, d$) & \textcolor{black!45}{0.231} & $-$2.379 & \textcolor{black!45}{0.260} & $-$1.869 & \textcolor{black!45}{0.368} & $-$1.225 & \textcolor{black!45}{0.178} & $-$0.187 &  &  \\
    + leverage \& ceiling ($\kappa, \lambda_{\mathrm{ceil}}$) & 0.224 & $-$2.414 & 0.231 & $-$1.894 & 0.290 & $-$1.270 & 0.112 & $-$1.225 & 0.271 & $-$1.251 \\
    \arrayrulecolor{black!25}\midrule\arrayrulecolor{black}
    + time term (\ours) & \textcolor{black!45}{0.224} & \textcolor{black!45}{$-$2.414} & \textcolor{black!45}{0.231} & \textcolor{black!45}{$-$1.894} & \textcolor{black!45}{0.290} & \textcolor{black!45}{$-$1.270} & \textcolor{black!45}{0.112} & \textcolor{black!45}{$-$1.225} & 0.234 & $-$1.359 \\
    \bottomrule
  \end{tabular}
\end{table*}

%% file: tables/swap.tex
\begin{table}[t]
  \centering
  \small
  \setlength{\tabcolsep}{2.8pt}
  \caption{Pairing an error level with a leverage on the sparse rig (N3DV, 4C4D), mean over six scenes (lower is
  better), without the ceiling. FisherRF's parameter uncertainty leaves either error level unchanged, whereas our leverage ($h$)
  improves both, most with the error level it is derived with, as in \ours.}
  \label{tab:swap}
  \begin{tabular}{@{}lcccccc@{}}
    \toprule
    & \multicolumn{2}{c}{no leverage} & \multicolumn{2}{c}{$\times$(1+$\kappa\,$FisherRF)}
    & \multicolumn{2}{c}{$\times$(1+$\kappa\,h$)} \\
    \cmidrule(lr){2-3}\cmidrule(lr){4-5}\cmidrule(l){6-7}
    Error level & AUSE & NLL & AUSE & NLL & AUSE & NLL \\
    \midrule
    3DGS-U & 0.358 & $-$1.237 & 0.357 & $-$1.238 & 0.321 & $-$1.258 \\
    \ours & 0.368 & $-$1.225 & 0.368 & $-$1.225 & \textbf{0.292} & \textbf{$-$1.263} \\
    \bottomrule
  \end{tabular}
\end{table}

%% file: sec/05_limitations.tex
\section{Limitations}
\label{sec:limitations}

\ours needs one camera withheld from training to fit the scales of its terms, and the estimate can be offset for views unlike that camera.
The leverage covers the appearance of the primitives, not their geometry, whose errors enter only through the
error level; a leverage of the geometry and motion parameters from their Fisher
information~\cite{jiang2024fisherrf,hanson2025pup} would cover them, but needs each backbone's motion Jacobians.
Low accumulated opacity does not always mean missing geometry, hence the fitted weight $\lambda_{\mathrm{ceil}}$; a per-pixel test
of whether training rays saw the background there could replace it.
Beyond the observed span, the time term assumes, as do dynamical-system
forecasters~\cite{asiimwe2025evogs}, that the scene keeps changing as it did within the span: where motion slows or
stops, it overestimates the error, and where motion starts, it underestimates it.
The linear extension is also pessimistic at long horizons (supplement); a growth fitted to the measured lags, or one that levels off, could avoid this.
We leave these extensions to future work.

%% file: sec/06_conclusion.tex
\section{Conclusion}
\label{sec:conclusion}

We presented \ours, a post-hoc estimator of the per-pixel variance of the rendering error of a trained dynamic
Gaussian splatting model at any viewpoint and time.
As in a regression prediction interval, it multiplies the error level fitted to the training residuals by one
plus a leverage computed with this level as noise; on a sparse rig, the gain comes from this leverage, and one built
from FisherRF's information leaves the estimate unchanged.
Against three post-hoc methods for static scenes adapted to these backbones, it achieves the best AUSE and NLL in
five settings, each with a different backbone family, and estimates a view's PSNR
with a mean absolute error of 0.61--2.03\,dB, so a free-viewpoint system can tell which views will
render poorly before showing them.

\section*{Acknowledgments}
This work was supported by the Australian Research Council Research Hub in Intelligent Robotic Systems for Real-Time
Asset Management (IH210100030) (ARIAM). We acknowledge support from the Queensland University of Technology (QUT) through its provision
of subsidised access to High Performance Computing. C.J., N.S., and D.M.\ also acknowledge ongoing support from
the QUT Centre for Robotics. T.G., P.S., and M.R.\ acknowledge support by the state of North Rhine-Westphalia and the
European Union within the EFRE/JTF project ``Just scan it 3D'', grant no.\ EFRE-20800529.

%% file: sec/XX_suppl.tex
\clearpage
\appendix
\setcounter{figure}{0}\setcounter{table}{0}\setcounter{equation}{0}
\renewcommand{\thefigure}{S\arabic{figure}}
\renewcommand{\thetable}{S\arabic{table}}
\renewcommand{\theequation}{S\arabic{equation}}
\section*{Supplementary Material}

In this supplement, we provide methodological details, describe the baselines, and report additional experiments. \Cref{sec:supp_method} expands on the method. Specifically, \cref{sec:supp_error_level} details the error-level model and its hyperparameters, while \cref{sec:supp_leverage} derives the posterior covariance of the color correction, introduces the per-primitive block approximation, bounds its overestimation, and explains its estimation using random signs. \Cref{sec:supp_calib} then describes the time term, ceiling estimation, and calibration procedure.

\Cref{sec:supp_baselines} describes the three baselines and their implementation, followed by additional results in \cref{sec:supp_results}. These include further metrics (\cref{sec:supp_metrics}), separate analyses of leverage and ceiling effects (\cref{sec:supp_lev_ceil}), monocular reconstruction on pixels also visible to the training camera (\cref{sec:supp_covisible}), more views (\cref{sec:supp_gallery}), and the sparse rig per group of test cameras, with views close to the rig and outside its span (\cref{sec:supp_groups}). We further distinguish moving and static pixels in dense captures (\cref{sec:supp_dynamic}), examine how rendering changes grow with temporal lag and a 75/25 split of the sequences (\cref{sec:supp_time}), detail experiments swapping components with the baselines (\cref{sec:supp_swap}), and report runtime and memory (\cref{sec:supp_runtime}).

\section{Method details}
\label{sec:supp_method}

\subsection{Error level}
\label{sec:supp_error_level}
\textbf{Basis.}
The log error level of primitive $k$ at camera $c$ and time $t$ is
$\eta_k(c,t)=\mathbf Y(\vomega_k)^\top\vbeta^{\vomega}_k+\mathbf B(t)^\top\mathbf P\,\vbeta^t_k$, with $\mathbf Y$ the
$(L{+}1)^2$ real SH up to degree $L$ and $\mathbf B$ the $M$ clamped cubic B-splines on $[0,1]$.
As the B-splines sum to one, part of the base level could be captured in $\vbeta^t_k$.
The projection $\mathbf P=\mathbf I-\mathbf 1\mathbf 1^\top/M$ removes it, so a primitive's base level is concentrated
in $\beta^{\vomega}_{k,0}$; this gives the shifted splines
$\tilde{\mathbf B}=\mathbf P\mathbf B=\mathbf B-\mathbf 1/M$ of \cref{sec:field}.

\textbf{Splatting.}
We splat the standard deviations of the primitives,
$s(p)=\sum_kw_k(p)\,e^{\eta_k/2}+\bigl(1-A(p)\bigr)\sqrt b$, and set $\eta(p)=2\log s(p)+g(t)$, where
$g(t)=\mathbf B(t)^\top\vbeta^g$ is the scene-global, time-dependent offset.
A blended pixel thus takes its standard deviation from its primitives in proportion to their weights, while compositing
$\eta_k$ itself would give the lower weighted geometric mean of their variances.
\Cref{tab:supp_blend} compares the two blends: blending the standard deviations gives the better AUSE and NLL in
all settings. The differences are small on N3DV and NVIDIA (at most 0.008 in AUSE and 0.007 in NLL) and large on
VRU, where blending the logarithms raises AUSE from 0.231 to 0.368.

\input{tables/supp_blend}

\textbf{Targets.}
We pool the squared residuals $y_0(p)=\frac13\lVert\hat I(p)-I(p)\rVert^2$ over covered pixels ($A(p)>\frac12$) with
a Gaussian kernel of standard deviation 2.5 pixels in an $11\times11$ window, renormalized over the covered pixels.
We fit only covered pixels, and calibration and evaluation use unpooled residuals.
We set $b$ to the 95th percentile of the targets.
Between their 1st and 99th percentiles, the targets of one scene span 2.1 to 3.7 orders of magnitude, over 22 scenes
of N3DV (dense and sparse rigs, future frames), VRU and NVIDIA.

\textbf{Prior.}
The coefficients $\vbeta_k=(\vbeta^{\vomega}_k;\vbeta^t_k)$ have precision
$\mPi=\blockdiag\bigl(\diag(\zeta_l^{-1}),\mathbf P^\top\mathbf K^{-1}\mathbf P\bigr)$ around the constant field $b$.
For directions, the heat kernel on the sphere with bandwidth $\varrho_{\vomega}$ has the SH as eigenfunctions
(Funk--Hecke~\cite[Thm.~1.2.9]{dai2013approximation}), so this block is diagonal with variance $\zeta_l=e^{-l(l+1)\varrho_{\vomega}}$ for degree
$l$~\cite{borovitskiy2020matern}.
For time, $\mathbf K$ is the Mat\'ern-3/2 covariance between the spline coefficients, evaluated at the Greville
abscissae of the splines, with length scale $\varrho_t$~\cite{rasmussen2006gaussian}.
$\mPi$ is singular only along the constant temporal direction; we invert it on its complement.

\textbf{Objective.} We minimize:
\begin{equation}
\begin{split}
  \mathcal L = \frac1{\lvert\mathcal P\rvert}&\sum_{p\in\mathcal P}\ell\bigl(\eta(p),y(p)\bigr)
  + \tau_g\lVert\vbeta^g\rVert^2 \\
  + &\sum_k (\vbeta_k-\vbeta^b)^\top\mPi_\tau\,(\vbeta_k-\vbeta^b),
\end{split}
\end{equation}
over sampled training pixels $\mathcal P$, with prior mean $\vbeta^b$, whose only nonzero entry is the degree-0 SH coefficient, set so that
$e^{\eta_k}=b$ at every direction and time, and
$\mPi_\tau=\blockdiag\bigl(\diag(\tau_0\zeta_0^{-1},\tau_{\vomega}\zeta_1^{-1},\ldots,\tau_{\vomega}\zeta_L^{-1}),
\tau_t\mathbf P^\top\mathbf K^{-1}\mathbf P\bigr)$.
The weights are small, so the prior decides only where data is scarce.
The minimization is stochastic with one rasterization per image and step.
We use $L=3$, $M=12$, $\varrho_{\vomega}=0.1$, $\varrho_t=0.15$ (in units of the observed span),
$\tau_0=3\cdot10^{-10}$, $\tau_{\vomega}=\tau_t=10^{-8}$ and $\tau_g=10^{-3}$, and 2000 Adam steps with a cosine
learning rate from 0.3 to 0.03, each on 65{,}536 target pixels from four training images.
We compute the targets on one random frame out of every four for each training camera, and on every frame for
monocular captures.
Apart from this, all values are shared across backbones and settings.

\subsection{Leverage}
\label{sec:supp_leverage}
The color correction from \cref{eq:color-correction} reads
\begin{equation}
    r(p) = \mathbf x(p)^\top \vtheta + \epsilon(p) = \sum_kw_k(p)\vpsi_k^\top\vtheta_k + \epsilon(p),
\end{equation}
with $\epsilon(p) \sim \mathcal N\bigl(0,e^{\eta(p)}\bigr)$.
Here and below, $p$ runs over the pixels of all training images $f=(c,t)$, and for a pixel of image $f$ we abbreviate
$w_k(p)=w_k(p;c,t)$, $\eta(p)=\eta(p;c,t)$ and $\vpsi_k=\vpsi_{kf}$, the features of primitive $k$ at the direction
and time of that image.
Since the model applies to each color channel independently, with identical weights, features, noise and prior,
the three posteriors have the same covariance, and we consider a single channel.
We assume the prior $\vtheta_k\sim\mathcal N(\mathbf 0,b\,\mPi^{-1})$ on the per-primitive parameters of the color correction $\vtheta$.
$\mPi$ is singular along the constant temporal direction, along which $\vpsi_k$ has no component, so neither the
correction nor the training residuals depend on this component of $\vtheta_k$.
We therefore restrict each $\vtheta_k$ to the complement of this direction, on which $\mPi$ is positive definite, and
take all inverses below on this complement.
The posterior covariance of $\vtheta$ is
\begin{equation}
    \mSigma=\bigl(\mathbf X^\top\mathbf W\mathbf X+\mathbf I_N\otimes\mPi/b\bigr)^{-1},
\end{equation}
where $\mathbf{X}$ has one row $\mathbf{x}(p)^\top$ per training pixel, $\mathbf{W} = \diag\bigl(e^{-\eta(p)}\bigr)$ is diagonal over the same pixels, $\mathbf{I}_N$ is the identity matrix with $N$ the number of primitives, and $\otimes$ is the Kronecker product. We now derive $\mSigma$.
The probability of the residuals $\mathbf r$ of all training pixels under $\vtheta$, assuming independence of
pixels, is:
\begin{equation}
    p(\mathbf r \vert \vtheta) = \prod_p \frac{1}{\sqrt{2\pi e^{\eta(p)}}}\exp\biggl(\!-\frac{(r(p)-\mathbf x(p)^\top \vtheta)^2}{2e^{\eta(p)}}\!\biggr),
    \label{eq:supp_prtheta}
\end{equation}
which gives the negative log-likelihood
\begin{align}
    -\log p(\mathbf r \vert \vtheta) =
    \frac12 (\mathbf r - \mathbf{X}\vtheta)^\top \mathbf{W} (\mathbf r- \mathbf{X}\vtheta) + \const,
\end{align}
where $\const$ is a constant independent of $\vtheta$.
The prior on $\vtheta$ is, up to a normalization that does not depend on $\vtheta$,
\begin{equation}
    p(\vtheta) \propto \prod_k \exp\biggl(-\frac12 \vtheta_k^\top \frac{\mPi}{b} \vtheta_k\biggr),
\end{equation}
with negative logarithm
\begin{equation}
\begin{split}
    -\log p(\vtheta) &= \frac12 \sum_k \vtheta_k^\top \frac{\mPi}{b}\vtheta_k + \const \\
    &= \frac12 \vtheta^\top \biggl(\mathbf{I}_N \otimes \frac{\mPi}{b} \biggr) \vtheta + \const.
\end{split}
\end{equation}
By Bayes' rule, the posterior $p(\vtheta \vert \mathbf r)$ is proportional to $p(\mathbf r \vert \vtheta) \cdot p(\vtheta)$, since $p(\mathbf r)$ does not depend on $\vtheta$.
With $\mathbf{A} := \mathbf{X}^\top \mathbf{W} \mathbf{X} + \mathbf{I}_N \otimes \mPi/b$, this gives:
\begin{equation}
\begin{split}
    -\log p(\vtheta \vert \mathbf r) &= - \log p(\mathbf r \vert \vtheta) - \log p(\vtheta) + \const \\
    &= \frac12 \vtheta^\top \mathbf{A}\,\vtheta - \vtheta^\top \mathbf{X}^\top \mathbf{W} \mathbf r + \const.
\end{split}
\end{equation}
This is quadratic in $\vtheta$, so the posterior is Gaussian. To read off its covariance, we complete the square with $\vmu := \mathbf{A}^{-1} \mathbf{X}^\top \mathbf{W} \mathbf r$, which yields
\begin{equation}
    - \log p(\vtheta \vert \mathbf r) = \frac12 (\vtheta- \vmu)^\top \mathbf{A}(\vtheta - \vmu) + \const.
\end{equation}
Thus $\vtheta \vert \mathbf r \sim \mathcal{N}(\vmu, \mathbf{A}^{-1})$, \ie the posterior is Gaussian with covariance $\mSigma = \mathbf{A}^{-1}$.

\textbf{Block structure.}
The variance of the correction at a query pixel $q$ is $\mathbf x(q)^\top\mSigma\,\mathbf x(q)$, which needs the
inverse of $\mathbf A$.
$\mathbf A$ consists of $N\times N$ blocks $\mathbf A_{kj}$ of size $D\times D$, one per pair of primitives, with
$D=(L+1)^2+M-1$ coefficients per primitive on the complement of the constant temporal direction.
To compute them, we write $\mathbf X^\top\mathbf W\mathbf X$ as a sum over the rows of $\mathbf X$:
\begin{equation}
  \mathbf X^\top\mathbf W\mathbf X=\sum_p e^{-\eta(p)}\,\mathbf x(p)\,\mathbf x(p)^\top.
  \label{eq:supp_outer}
\end{equation}
For a pixel $p$ of image $f$, $\mathbf x(p)$ stacks the vectors $w_k(p)\,\vpsi_{kf}$ of all primitives, so the block
of $\mathbf x(p)\,\mathbf x(p)^\top$ for primitives $k$ and $j$ is $w_k(p)\,w_j(p)\,\vpsi_{kf}\vpsi_{jf}^\top$.
Collecting the pixels $p\in f$ of each image gives the block of $\mathbf X^\top\mathbf W\mathbf X$,
\begin{equation}
\begin{gathered}
  \bigl(\mathbf X^\top\mathbf W\mathbf X\bigr)_{kj}=\sum_f S_{kjf}\,\vpsi_{kf}\vpsi_{jf}^\top,\\
  S_{kjf}=\sum_{p\in f}w_k(p)\,w_j(p)\,e^{-\eta(p)}.
\end{gathered}
\label{eq:supp_blocks}
\end{equation}
The prior term $\mathbf I_N\otimes\mPi/b$ adds $\mPi/b$ to each diagonal block and nothing to the others, so
$\mathbf A_{kj}$ is the block of \cref{eq:supp_blocks} for $k\neq j$, and that block plus $\mPi/b$ for $k=j$.
An off-diagonal block is nonzero as soon as two primitives share one training pixel, so $\mathbf A$ couples all
primitives along every viewing ray, and its $DN\times DN$ inverse cannot be computed for a full scene.

\textbf{Block approximation.}
We therefore keep only the diagonal blocks.
For $k=j$, \cref{eq:supp_blocks} and the prior term give
\begin{equation}
\begin{gathered}
  \mathbf A_{kk}=\mathbf H_k+\mPi/b,\\
  \mathbf H_k=\sum_f\Bigl(\sum_{p\in f}w_k(p)^2\,e^{-\eta(p)}\Bigr)\vpsi_{kf}\vpsi_{kf}^\top,
\end{gathered}
\end{equation}
with $\mathbf H_k$ the Fisher block of \cref{sec:leverage}.
We replace $\mathbf A$ by its block diagonal $\tilde{\mathbf A}=\blockdiag(\mathbf A_{11},\ldots,\mathbf A_{NN})$,
whose inverse is $\blockdiag(\mSigma_1,\ldots,\mSigma_N)$ with $\mSigma_k=(\mathbf H_k+\mPi/b)^{-1}$.
As $\mathbf x(q)$ stacks the vectors $w_k(q)\,\vpsi_k$, with $\vpsi_k$ at the direction and time of the query, the
query variance becomes
\begin{equation}
  \mathbf x(q)^\top\tilde{\mathbf A}^{-1}\mathbf x(q)=\sum_kw_k(q)^2\,\vpsi_k^\top\mSigma_k\vpsi_k=v(q),
  \label{eq:supp_vq}
\end{equation}
which is \cref{eq:blocks}.

\textbf{Error of the block approximation.}
The blocks drop the posterior correlations between primitives that share a training pixel, which observes only
the blend of their colors, not each color on its own.
The variance of a blend seen in training is thereby overestimated, by a bounded factor; that of a blend never seen
can be underestimated without bound.
Let $m$ be the largest number of primitives with nonzero weight at one training pixel.
For any coefficient vector $\vtheta$ and training pixel $p$,
$\mathbf x(p)^\top\vtheta=\sum_kw_k(p)\,\vpsi_k^\top\vtheta_k$ has at most $m$ nonzero terms, so by the
Cauchy--Schwarz inequality
\begin{equation}
  \bigl(\mathbf x(p)^\top\vtheta\bigr)^2\le m\sum_k\bigl(w_k(p)\,\vpsi_k^\top\vtheta_k\bigr)^2.
\end{equation}
Multiplying by $e^{-\eta(p)}$ and summing over $p$ gives, with \cref{eq:supp_outer},
$\vtheta^\top\mathbf X^\top\mathbf W\mathbf X\,\vtheta\le m\sum_k\vtheta_k^\top\mathbf H_k\vtheta_k$.
The prior term is the same in $\mathbf A$ and $\tilde{\mathbf A}$, and $m\ge1$, so
$\vtheta^\top\mathbf A\,\vtheta\le m\,\vtheta^\top\tilde{\mathbf A}\,\vtheta$ for every $\vtheta$, \ie
$\mathbf A\preceq m\,\tilde{\mathbf A}$.
Inversion reverses this order between positive definite matrices, so
$\tilde{\mathbf A}^{-1}\preceq m\,\mathbf A^{-1}$, and for every query pixel
\begin{equation}
  v(q)\le m\;\mathbf x(q)^\top\mathbf A^{-1}\mathbf x(q).
\end{equation}
The blocks thus overestimate the variance of the full model by at most the factor $m$.
This bound is tight.
Consider one coefficient per primitive and a query that blends $m$ primitives with the same weights as $n$ training
pixels at the error level $e^{\eta}$.
These pixels determine only the blend, not the individual coefficients, so as $n$ grows the full variance decays as
$e^{\eta}/n$, whereas the block variance decays as $m\,e^{\eta}/n$, and their ratio tends to $m$.
No such bound holds in the other direction.
At a query that separates primitives always seen together, \eg after an occlusion change, the full variance remains
on the order of the prior level, because the training pixels never distinguished these primitives.
The blocks, in contrast, credit every such pixel to each primitive separately and therefore underestimate the
variance.

A second approximation is shared with the full model: treating the correlated pixels of a camera, across neighbors
and frames, as independent (\cref{eq:supp_prtheta}) overcounts the evidence and shrinks $h$ where the data dominate.
The calibrated scale $\kappa$ absorbs the typical size of both effects: since
$\kappa(\mathbf H_k+\mPi/b)^{-1}=\bigl(\mathbf H_k/\kappa+\mPi/(\kappa b)\bigr)^{-1}$, $n$ correlated pixels count as $n/\kappa$
where the data dominate, as with the design effect of clustered samples~\cite{kish1965survey}.

\textbf{Estimation with random signs.}
The inner sum of $\mathbf H_k$, $S_{kf}=S_{kkf}=\sum_pw_k(p;c,t)^2e^{-\eta(p;c,t)}$ for training image $f=(c,t)$, and $v(q)=\textstyle\sum_k w_k(q)^2\,v_k$
are sums of squared weights, but the rasterizer does not expose $w_k$ or $w_k^2$.
Random signs solve both: for independent signs $\xi_i\in\{\pm1\}$ and any coefficients $a_i$,
\begin{equation}
  \mathbb E\bigl[(\textstyle\sum_ia_i\xi_i)^2\bigr]=\textstyle\sum_ia_i^2,
  \label{eq:supp_signs}
\end{equation}
as the cross terms vanish in expectation.
For $S_{kf}$, the signs go on the pixels.
We assign every primitive a scalar and render these scalars from $f$.
Backpropagating a Rademacher image $\xi(p)$, scaled by $e^{-\eta(p;c,t)/2}$, through this render gives the gradient
with respect to the scalar of primitive $k$,
\begin{equation}
  u_k=\textstyle\sum_pw_k(p;c,t)\,e^{-\eta(p;c,t)/2}\,\xi(p),
  \label{eq:supp_probe}
\end{equation}
with $\mathbb E_\xi[u_k^2]=S_{kf}$ by \cref{eq:supp_signs}, the randomized Gauss--Newton diagonal
estimator~\cite{martens2012training}; independent probes per color channel and the sum over training images reduce
its noise.
For $v(q)$, the signs go on the primitives: we splat $\sqrt{v_k}\,z_k$ with $z_k\in\{\pm1\}$ and square the image,
which equals $v(q)$ in expectation~\cite{alon1996space,hutchinson1989stochastic}.
We average several such renders to $\bar v$, pool it like the targets of the error level, and set
$h(q)=\bar v(q)\,e^{-\eta(q)}$.
We use three probes for $S_{kf}$, one per color channel, on every training image, and 24 for $v$.

\input{tables/supp_metrics}

\subsection{Time term, ceiling and calibration}
\label{sec:supp_calib}
\textbf{Time term.}
We measure $\gamma_k(l)$ at five lags, $l\in\{\frac1{32},\frac1{16},\frac18,\frac3{16},\frac14\}$ of the observed
span.
For each lag and each training camera $c$, we render all pairs of times $(s,s+l)$ within $[0,1]$, with $s$ on a grid
of step $\frac1{32}$, and take the squared change $\Delta^2_{csl}(p)$ of the render between the two times, averaged
over the color channels.
We gather this change onto the primitives with the weights at $s$:
\begin{equation}
  \gamma_k(l)=\frac{\sum_{c,s}\sum_pw_k(p;c,s)\,\Delta^2_{csl}(p)+\bar\Delta^2_l}
  {\sum_{c,s}\sum_pw_k(p;c,s)+1}.
\end{equation}
The added terms act as one more pixel at the mean change $\bar\Delta^2_l$ of all pairs, which pulls barely seen
primitives toward this mean.
For a horizon $\delta\leq\frac14$, we interpolate $\gamma_k$ linearly between the neighboring lags,
\begin{equation}
    \gamma_k(\delta)= \gamma_k(l_-) + \frac{\gamma_k(l_+) - \gamma_k(l_-)}{l_+ - l_-}(\delta - l_-),
\end{equation}
with $l_-$ the largest lag not exceeding $\delta$ and $l_+$ the smallest lag above it; below the first lag,
$l_-=0$ and $\gamma_k(0)=0$.
Beyond the observed lags ($\delta>\frac14$), we assume linear growth with the horizon, $\gamma_k(\delta)=4\delta\,\gamma_k(\frac14)$, capped by the error ceiling.
As the measured change grows less than linearly, this extension overestimates the change at long horizons
(\cref{sec:supp_time}).

\textbf{Ceiling.}
The per-channel mean $\bar{\mathbf c}$ and variance $\vnu$ of the scene's colors are estimated from a subset of the
training images, which we choose in two steps.
First, we divide the frames of each training camera into consecutive groups of 30 and draw one frame at random from
each group, which spreads the samples evenly over time.
Second, we list these frames camera by camera and keep every fifth of the list.

\textbf{Calibration.}
We fit the four scalars in two steps on the residuals of the held-out camera, with the three color channels of a
pixel sharing one variance.
First, we fit $a$, $d$ and $\kappa$ with L-BFGS, minimizing the Gaussian negative log-likelihood of the uncapped
estimate $e^{a\eta+d}(1+\kappa h)$ on up to 20{,}000 random pixels per frame, from every second frame of the held-out
camera.
These frames lie within the span, where the time term is zero.
Second, we choose $\hat\lambda_{\mathrm{ceil}}$ on a grid of step 0.05 over $[0,1]$, minimizing the same likelihood for
the full estimate (\cref{eq:final}) on the same pixels.
We fit $\hat\lambda_{\mathrm{ceil}}$ because low opacity does not translate to missing geometry on every backbone.

\begin{figure*}[t]
  \centering
  \includegraphics[width=0.86\textwidth]{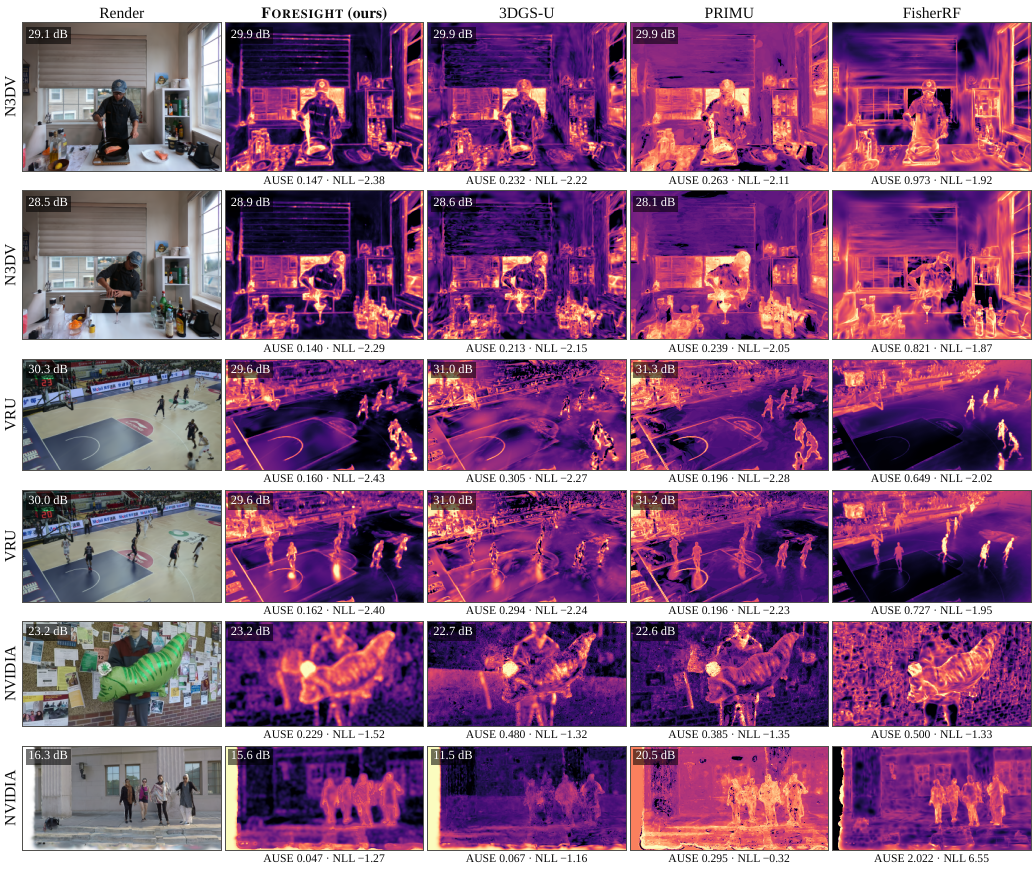}
  \caption{Additional visualizations on N3DV, VRU and NVIDIA. Labels give the true
  PSNR (on the render) and each method's estimate (none for FisherRF); below each map are the view's AUSE and NLL
  (lower is better).}
  \label{fig:supp_gallery}
\end{figure*}

\section{Baselines}
\label{sec:supp_baselines}

\textbf{Shared protocol.}
For a fair comparison, all baselines run on the same frozen backbone as \ours, with its primitives deformed to the
time of each image, and use only its training images.
\mbox{3DGS-U} and PRIMU learn from the squared rendering error of these images, as \ours does; FisherRF has no error
target, as it measures parameter uncertainty.
The logarithm $\eta$ of each baseline's per-pixel map gives the variance $e^{a\eta+d}$ and the PSNR estimate as for \ours,
with $a$ and $d$ fitted on the same pixels of the same held-out camera (\cref{sec:supp_calib}).

\textbf{\mbox{3DGS-U}.}
\mbox{3DGS-U}~\cite{galappaththige2026predictive} fits one additional SH channel per primitive, rendered like a color,
by least squares to a per-pixel error score of the training images.
We run its released code with two changes, for a fair comparison.
To give it the same error target as \ours, the channel is fitted to the square root of our targets $y$
(\cref{sec:field}) instead of its $L_1$ and D-SSIM score.
To give it the same image budget as \ours, it runs for 8{,}000 iterations of one image each instead of 3{,}000.
The released channel depends on the viewing direction but not on time.
Conditioning it on time, as the backbone conditions color, slightly improves its AUSE on the dense rig of N3DV (0.246 \vs 0.251) and
worsens it on the sparse rig (0.379 \vs 0.358, in all six scenes), with NLL within 0.01; on Shape of Motion, whose colors are constant
in time, the conditioned channel is the released one. We therefore keep the release.
The rendered value $u$ enters the calibration as $\eta=\log(u+1/255)$.

\textbf{PRIMU.}
We follow the release of PRIMU~\cite{gottwald2026primu}, with two changes to its gradient-boosted regressor, for a
fair comparison.
To give it the same training information as \ours and the other baselines, we fit the regressor on training images;
the held-out camera then only sets the shared calibration, as for every method.
To give it the same error target as \ours, the regressor fits the per-pixel squared error instead of the $L_1$ error.
Its prediction $\hat y$ enters the calibration as $\eta=\log\max(\hat y,10^{-6})$.

\textbf{FisherRF.}
We follow the release of FisherRF~\cite{jiang2024fisherrf}, with its Fisher information taken over the color
parameters of each backbone: SH coefficients, the color logits of Shape of Motion and ATGS, or the 4D coefficients of
4C4D.
To give it the same training information as \ours and the other baselines, we sum the information over the training
images; the held-out camera only sets the shared calibration, as for every method.

\section{Additional results}
\label{sec:supp_results}

\subsection{Further metrics}
\label{sec:supp_metrics}
\Cref{tab:supp_metrics} reports two further metrics for the settings of \cref{tab:main,tab:sparse,tab:time}: the
Spearman correlation $\rho$ of the estimated variance with the squared error, and the expected calibration error (ECE)
of the Gaussian intervals~\cite{kuleshov2018accurate}.
\ours has the highest $\rho$ in all five settings and the lowest ECE in four.
On future frames, the ECE of \mbox{3DGS-U} is slightly lower (0.043 \vs 0.051): the intervals of \ours cover more than their
nominal level there, as the linear extension of the time term is pessimistic (\cref{sec:limitations}).

\subsection{Leverage and ceiling}
\label{sec:supp_lev_ceil}
\Cref{tab:ablation} adds the leverage and the ceiling in one step, as the ceiling bounds the estimate where the
leverage grows without limit.
\Cref{tab:supp_ablation} takes them apart on the same runs.

\input{tables/supp_ablation}

\subsection{Monocular capture on covisible pixels}
\label{sec:supp_covisible}
\Cref{tab:supp_nvidia_cov} repeats the monocular results of \cref{tab:main} on the pixels the training camera also
sees.
There, \mbox{3DGS-U}, PRIMU and \ours estimate the PSNR to within 1.1 to 1.3\,dB; the gap on the full frame comes
from the pixels the training camera never observed.
On these pixels, \ours keeps the best AUSE, NLL, $\rho$ and ECE.

\input{tables/supp_nvidia_cov}

\subsection{More views}
\label{sec:supp_gallery}
\Cref{fig:supp_gallery} shows two more test views in each setting of \cref{fig:qual}.

\subsection{Sparse rig: camera groups and views outside the span}
\label{sec:supp_groups}
4C4D trains on cam01, cam10, cam13 and cam20, and we calibrate on cam18.
We group the other held-out cameras present in all six scenes by their poses in the first frame, in the units of the
released camera poses of 4C4D.
A camera is \emph{close} if its center lies less than 0.6 from the nearest training camera (cam02, cam09, cam14 and
cam19), and \emph{outside} the span of the training cameras if it lies more than 0.1 outside the convex hull of their
centers in the plane of the rig (cam11 and cam12); all others lie \emph{between} them (cam00, cam05, cam06, cam07, cam08
and cam16).
\ours has the best AUSE and NLL in each group (\cref{tab:supp_sparse_groups}).
Its $\rho$ is second to that of \mbox{3DGS-U} on the close cameras (0.300 \vs 0.305), and its ECE second to that of
FisherRF on the cameras between (0.060 \vs 0.059).

\input{tables/supp_sparse_groups}

\begin{figure}[t]
  \centering
  \includegraphics[width=\columnwidth]{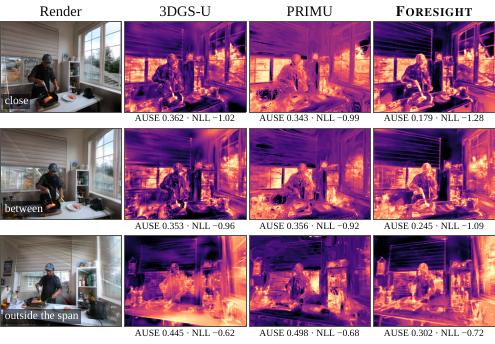}
  \vspace{-1.4em}
  \caption{Sparse rig (N3DV, 4C4D): a close camera, one between the training cameras and one outside their span
  (\cref{sec:supp_groups}) at the same frame, with the uncertainty of \mbox{3DGS-U}, PRIMU and \ours. Below each map are the view's
  AUSE and NLL (lower is better).}
  \label{fig:sparse_ood}
\end{figure}

\Cref{fig:sparse_ood} shows a close camera, one between and one outside the span of one scene at the same frame, with the
uncertainty of \mbox{3DGS-U}, PRIMU and \ours.
\label{sec:supp_sparse}

\subsection{Moving and static pixels on the dense rigs}
\label{sec:supp_dynamic}
We split the pixels of each test view by the temporal standard deviation of the ground-truth luminance at that pixel
over the test camera's video: above 0.05, the pixel counts as moving.
This marks 6\% of the pixels as moving on N3DV and 34\% on VRU.
\ours has the best AUSE and NLL on the moving and on the static pixels of both datasets (\cref{tab:supp_motion}).
On the moving pixels, it lowers the AUSE of the best baseline by 18\% on N3DV (0.162 \vs 0.198 for \mbox{3DGS-U}) and
by 27\% on VRU (0.213 \vs 0.290 for PRIMU).

\input{tables/supp_motion}

\subsection{Future frames: the time term}
\label{sec:supp_time}
\textbf{Backbone.}
Of the dynamic Gaussian splatting methods that forecast future frames, EvoGS~\cite{asiimwe2025evogs} is, to our
knowledge, the only one evaluated on multi-view capture such as N3DV; the others are evaluated on synthetic scenes and
handheld captures (\cref{sec:related_work}).
Its forecasts are less accurate than its renders within the observed span (PSNR of cam00 22.2 \vs 25.3\,dB, mean over
the six scenes), so future frames test the uncertainty under larger rendering errors.
As a post-hoc method, \ours applies unchanged to stronger forecasting backbones as they appear.

\textbf{Growth with the lag.}
Over the measured lags, from $\frac1{32}$ to $\frac14$ of the span, an eightfold range, the mean change
$\bar\Delta^2_l$ grows by a factor of 1.7 to 5.0 on N3DV (EvoGS and 4DGaussians), less than the eightfold growth
of the linear extension.
Extended linearly beyond $\frac14$, the time term therefore overestimates the change at long horizons.
On the future half of each sequence, about three quarters of the frames lie beyond this lag, and with the 75/25 split
of EvoGS~\cite{asiimwe2025evogs}, about a quarter.

\textbf{The 75/25 split.}
\Cref{tab:time} uses the default of the official EvoGS implementation, which observes the first half of each sequence.
\Cref{tab:supp_time75} repeats it with the 75/25 split of the EvoGS paper: EvoGS observes frames 0 to 225, with static
anchors at frames 0, 75, 150 and 225, and cam00 is scored on frames 226 to 299; all methods are calibrated on the
held-out camera within the observed span, as before.
\ours has the best AUSE, NLL and $\Delta$PSNR, as on the half split.

\input{tables/supp_time75}

\subsection{Swapping parts with the baselines}
\label{sec:supp_swap}
In \cref{tab:swap}, the error level $\eta$ is that of \ours or the channel of \mbox{3DGS-U}, and each is paired with no leverage, with ours, or with a leverage of the same form
built from FisherRF's per-primitive color information: $v_k=(H_k/\bar s^2+1/b)^{-1}$, with $H_k$ FisherRF's Fisher
information of the color of primitive $k$, its constant noise $\bar s^2$ set to the mean of our targets, and our prior
level $b$.
As ours, $v_k$ is composited with squared weights and smoothed, and $h=v\,e^{-\eta}$ uses the cell's own error level; our
leverage keeps our error level as its noise in both rows.
Every cell is calibrated as \ours ($a$, $d$ and $\kappa$; $a$ and $d$ without a leverage) and is not capped at the
ceiling: the cell of \mbox{3DGS-U} without a leverage reproduces its row of \cref{tab:sparse}, and that of \ours with our
leverage differs from it only by the ceiling (AUSE 0.292 \vs 0.290).

\subsection{Runtime and memory}
\label{sec:supp_runtime}
\Cref{tab:supp_runtime} reports the runtime and the peak GPU memory of every method on one N3DV scene.
After the 20 minutes of training of the backbone, \ours takes 6.2 minutes per scene, mostly to fit the error level and to
accumulate the Fisher blocks of the leverage, and 0.06\,s per test frame.
Its peak GPU memory, 3.6\,GB, is less than half that of training the backbone.

\input{tables/supp_runtime}

%% file: tables/supp_blend.tex
\begin{table}[t]
  \centering
  \small
  \setlength{\tabcolsep}{3pt}
  \caption{Blending the standard deviations (SD) $e^{\eta_k/2}$ of the primitives into the error level of a pixel, as in
  \ours (\cref{eq:eta}), against blending their logarithms $\eta_k$.
  Spearman correlation $\rho$ of the estimate with the squared error over pixels.
  The better value is in bold.}
  \label{tab:supp_blend}
  \resizebox{\columnwidth}{!}{%
  \begin{tabular}{@{}l*{6}{c}@{}}
    \toprule
    & \multicolumn{3}{c}{SD (\ours)} & \multicolumn{3}{c}{Logarithms} \\
    \cmidrule(lr){2-4} \cmidrule(l){5-7}
    Setting & $\rho\uparrow$ & AUSE$\downarrow$ & NLL$\downarrow$ & $\rho\uparrow$ & AUSE$\downarrow$ & NLL$\downarrow$ \\
    \midrule
    N3DV (dense) & \textbf{0.324} & \textbf{0.224} & \textbf{$-$2.414} & 0.314 & 0.227 & $-$2.408 \\
    VRU (dense) & \textbf{0.463} & \textbf{0.231} & \textbf{$-$1.894} & 0.325 & 0.368 & $-$1.731 \\
    N3DV (sparse) & \textbf{0.322} & \textbf{0.290} & \textbf{$-$1.270} & 0.317 & 0.298 & $-$1.264 \\
    NVIDIA (monocular) & \textbf{0.514} & \textbf{0.112} & \textbf{$-$1.225} & 0.509 & 0.114 & $-$1.218 \\
    \bottomrule
  \end{tabular}}
\end{table}

%% file: tables/supp_metrics.tex
\begin{table*}[t]
  \centering
  \small
  \setlength{\tabcolsep}{5pt}
  \caption{Ranking and calibration on the settings of \cref{tab:main,tab:sparse,tab:time}: Spearman correlation $\rho$
  of the estimated variance with the squared error, and expected calibration error (ECE) of the Gaussian
  intervals~\cite{kuleshov2018accurate}. \ours has the \firstbox{best} $\rho$ in all settings and the best ECE in
  all but future frames; \secondbox{second best} is also highlighted.}
  \label{tab:supp_metrics}
  \begin{tabular}{@{}l*{10}{c}}  %
    \toprule
    & \multicolumn{2}{c}{N3DV (dense)} & \multicolumn{2}{c}{VRU (dense)} & \multicolumn{2}{c}{NVIDIA (monocular)} & \multicolumn{2}{c}{N3DV (sparse)} & \multicolumn{2}{c}{N3DV (future)} \\
    \cmidrule(lr){2-3} \cmidrule(lr){4-5} \cmidrule(lr){6-7} \cmidrule(lr){8-9} \cmidrule(l){10-11}
    Method & $\rho\uparrow$ & ECE$\downarrow$ & $\rho\uparrow$ & ECE$\downarrow$ & $\rho\uparrow$ & ECE$\downarrow$ & $\rho\uparrow$ & ECE$\downarrow$ & $\rho\uparrow$ & ECE$\downarrow$ \\
    \midrule
    FisherRF~\cite{jiang2024fisherrf} & 0.106 & 0.057 & 0.142 & 0.052 & 0.102 & 0.082 & 0.133 & \second{0.055} & 0.075 & 0.082 \\
    3DGS-U~\cite{galappaththige2026predictive} & \second{0.316} & \second{0.055} & \second{0.387} & 0.048 & \second{0.414} & \second{0.066} & \second{0.304} & 0.065 & \second{0.394} & \first{0.043} \\
    PRIMU~\cite{gottwald2026primu} & 0.221 & 0.059 & 0.374 & \second{0.044} & 0.307 & 0.081 & 0.264 & 0.065 & 0.279 & 0.069 \\
    \arrayrulecolor{black!25}\midrule\arrayrulecolor{black}
    \ours (ours) & \first{0.324} & \first{0.027} & \first{0.463} & \first{0.038} & \first{0.514} & \first{0.050} & \first{0.322} & \first{0.049} & \first{0.403} & \second{0.051} \\
    \bottomrule
  \end{tabular}
\end{table*}

%% file: tables/supp_ablation.tex
\begin{table*}[t]
  \centering
  \small
  \setlength{\tabcolsep}{5pt}
  \caption{\cref{tab:ablation} with the leverage and the ceiling taken apart.
  ``+ leverage'' omits the ceiling; ``+ ceiling, no leverage'' caps the calibrated error level ($\kappa=0$) at the
  ceiling, with its own fitted $\hat\lambda_{\mathrm{ceil}}$.
  On the rigs, the leverage carries the gain and the ceiling alone changes almost nothing.
  On monocular capture, the ceiling alone gives most of the gain, and the leverage, which needs the ceiling (without
  it, NLL diverges on pixels without reconstructed geometry), lowers AUSE by a further 15\%.
  Gray: identical to the row above by construction, as calibration is monotone.}
  \label{tab:supp_ablation}
  \begin{tabular}{@{}l*{8}{c}@{}}
    \toprule
    & \multicolumn{2}{c}{N3DV (dense)} & \multicolumn{2}{c}{VRU (dense)} & \multicolumn{2}{c}{N3DV (sparse)} & \multicolumn{2}{c}{NVIDIA (monocular)} \\
    \cmidrule(lr){2-3} \cmidrule(lr){4-5} \cmidrule(lr){6-7} \cmidrule(l){8-9}
    Component & AUSE$\downarrow$ & NLL$\downarrow$ & AUSE$\downarrow$ & NLL$\downarrow$ & AUSE$\downarrow$ & NLL$\downarrow$ & AUSE$\downarrow$ & NLL$\downarrow$ \\
    \midrule
    Error level & 0.231 & $-$1.468 & 0.260 & $-$1.218 & 0.368 & 17.462 & 0.178 & 31.041 \\
    + calibration ($a, d$) & \textcolor{black!45}{0.231} & $-$2.379 & \textcolor{black!45}{0.260} & $-$1.869 & \textcolor{black!45}{0.368} & $-$1.225 & \textcolor{black!45}{0.178} & $-$0.187 \\
    + leverage ($\kappa$) & 0.224 & $-$2.414 & 0.231 & $-$1.888 & 0.292 & $-$1.263 & 0.636 & $>\!10^{6}$ \\
    + ceiling ($\hat\lambda_{\mathrm{ceil}}$), no leverage & 0.231 & $-$2.379 & 0.260 & $-$1.871 & 0.368 & $-$1.226 & 0.131 & $-$1.178 \\
    \arrayrulecolor{black!25}\midrule\arrayrulecolor{black}
    + leverage \& ceiling (\ours) & 0.224 & $-$2.414 & 0.231 & $-$1.894 & 0.290 & $-$1.270 & 0.112 & $-$1.225 \\
    \bottomrule
  \end{tabular}
\end{table*}

%% file: tables/supp_nvidia_cov.tex
\begin{table*}[t]
  \centering
  \small
  \setlength{\tabcolsep}{5pt}
  \caption{Monocular capture (NVIDIA) on the full frame and on the pixels the training camera also sees; $\rho$ and ECE
  as in \cref{tab:supp_metrics}.
  \firstbox{Best} and \secondbox{second best} are highlighted; $\Delta$PSNR is not reported for FisherRF
  (\cref{tab:main}).}
  \label{tab:supp_nvidia_cov}
  \begin{tabular}{@{}l*{10}{c}}  %
    \toprule
    & \multicolumn{5}{c}{Full frame (\cref{tab:main})} & \multicolumn{5}{c}{Covisible pixels} \\
    \cmidrule(lr){2-6} \cmidrule(l){7-11}
    Method & AUSE$\downarrow$ & NLL$\downarrow$ & $\Delta$PSNR$\downarrow$ & $\rho\uparrow$ & ECE$\downarrow$ & AUSE$\downarrow$ & NLL$\downarrow$ & $\Delta$PSNR$\downarrow$ & $\rho\uparrow$ & ECE$\downarrow$ \\
    \midrule
    FisherRF~\cite{jiang2024fisherrf} & 1.504 & 4.794 & -- & 0.102 & 0.082 & 0.641 & $-$1.283 & -- & 0.117 & 0.073 \\
    3DGS-U~\cite{galappaththige2026predictive} & \second{0.161} & \second{$-$1.065} & \second{4.79} & \second{0.414} & \second{0.066} & \second{0.459} & \second{$-$1.329} & \first{1.13} & 0.252 & \second{0.070} \\
    PRIMU~\cite{gottwald2026primu} & 0.626 & 1.669 & 6.76 & 0.307 & 0.081 & 0.470 & $-$1.313 & 1.30 & \second{0.338} & 0.071 \\
    \arrayrulecolor{black!25}\midrule\arrayrulecolor{black}
    \ours (ours) & \first{0.112} & \first{$-$1.225} & \first{0.61} & \first{0.514} & \first{0.050} & \first{0.326} & \first{$-$1.431} & \second{1.18} & \first{0.388} & \first{0.059} \\
    \bottomrule
  \end{tabular}
\end{table*}

%% file: tables/supp_sparse_groups.tex
\begin{table*}[t]
  \centering
  \small
  \setlength{\tabcolsep}{5pt}
  \caption{Sparse rig (\cref{tab:sparse}) per group of test cameras (\cref{sec:supp_groups}); $\rho$ and ECE as in
  \cref{tab:supp_metrics}. \firstbox{Best} and \secondbox{second best} are highlighted.}
  \label{tab:supp_sparse_groups}
  \begin{tabular}{@{}l*{12}{c}}  %
    \toprule
    & \multicolumn{4}{c}{Close} & \multicolumn{4}{c}{Between} & \multicolumn{4}{c}{Outside} \\
    \cmidrule(lr){2-5} \cmidrule(lr){6-9} \cmidrule(l){10-13}
    Method & AUSE$\downarrow$ & NLL$\downarrow$ & $\rho\uparrow$ & ECE$\downarrow$ & AUSE$\downarrow$ & NLL$\downarrow$ & $\rho\uparrow$ & ECE$\downarrow$ & AUSE$\downarrow$ & NLL$\downarrow$ & $\rho\uparrow$ & ECE$\downarrow$ \\
    \midrule
    FisherRF~\cite{jiang2024fisherrf} & 0.550 & $-$1.294 & 0.103 & \second{0.078} & 0.458 & $-$1.202 & 0.155 & \first{0.059} & 0.472 & $-$0.910 & 0.127 & \second{0.031} \\
    3DGS-U~\cite{galappaththige2026predictive} & \second{0.404} & \second{$-$1.380} & \first{0.305} & 0.079 & \second{0.332} & \second{$-$1.249} & \second{0.328} & 0.061 & \second{0.344} & $-$0.915 & \second{0.231} & 0.079 \\
    PRIMU~\cite{gottwald2026primu} & 0.473 & $-$1.289 & 0.239 & 0.087 & 0.374 & $-$1.184 & 0.292 & 0.067 & 0.372 & \second{$-$0.919} & 0.230 & 0.037 \\
    \arrayrulecolor{black!25}\midrule\arrayrulecolor{black}
    \ours (ours) & \first{0.325} & \first{$-$1.407} & \second{0.300} & \first{0.053} & \first{0.267} & \first{$-$1.273} & \first{0.346} & \second{0.060} & \first{0.286} & \first{$-$0.984} & \first{0.292} & \first{0.029} \\
    \bottomrule
  \end{tabular}
\end{table*}

%% file: tables/supp_motion.tex
\begin{table}[t]
  \centering
  \small
  \setlength{\tabcolsep}{4pt}
  \caption{Dense rigs (\cref{tab:main}) on moving and static pixels: a pixel moves if the temporal standard deviation
  of the ground-truth luminance at it, over the test camera's video, exceeds 0.05. \firstbox{Best} and \secondbox{second best} are highlighted per dataset.}
  \label{tab:supp_motion}
  \begin{tabular}{@{}l*{4}{c}}  %
    \toprule
    & \multicolumn{2}{c}{Moving pixels} & \multicolumn{2}{c}{Static pixels} \\
    \cmidrule(lr){2-3} \cmidrule(l){4-5}
    Method & AUSE$\downarrow$ & NLL$\downarrow$ & AUSE$\downarrow$ & NLL$\downarrow$ \\
    \midrule
    \multicolumn{5}{@{}l}{\textit{N3DV}} \\
    FisherRF~\cite{jiang2024fisherrf} & 0.697 & $-$1.567 & 0.594 & $-$2.229 \\
    3DGS-U~\cite{galappaththige2026predictive} & \second{0.198} & \second{$-$2.006} & \second{0.269} & \second{$-$2.373} \\
    PRIMU~\cite{gottwald2026primu} & 0.361 & $-$1.687 & 0.330 & $-$2.305 \\
    \arrayrulecolor{black!25}\midrule\arrayrulecolor{black}
    \ours (ours) & \first{0.162} & \first{$-$2.063} & \first{0.242} & \first{$-$2.436} \\
    \midrule
    \multicolumn{5}{@{}l}{\textit{VRU}} \\
    FisherRF~\cite{jiang2024fisherrf} & 0.512 & $-$1.691 & 0.610 & $-$1.589 \\
    3DGS-U~\cite{galappaththige2026predictive} & 0.317 & \second{$-$1.880} & \second{0.309} & \second{$-$1.820} \\
    PRIMU~\cite{gottwald2026primu} & \second{0.290} & $-$1.799 & 0.349 & $-$1.569 \\
    \arrayrulecolor{black!25}\midrule\arrayrulecolor{black}
    \ours (ours) & \first{0.213} & \first{$-$2.001} & \first{0.233} & \first{$-$1.855} \\
    \bottomrule
  \end{tabular}
\end{table}

%% file: tables/supp_time75.tex
\begin{table}[t]
  \centering
  \small
  \setlength{\tabcolsep}{3pt}
  \caption{Future frames as in \cref{tab:time}, with EvoGS trained on the first three quarters of each sequence and
  evaluated on the last quarter.
  \firstbox{Best} and \secondbox{second best} are highlighted.}
  \label{tab:supp_time75}
  \begin{tabular}{@{}l*{3}{c}}  %
    \toprule
    Method & AUSE$\downarrow$ & NLL$\downarrow$ & $\Delta$PSNR$\downarrow$ \\
    \midrule
    FisherRF~\cite{jiang2024fisherrf} & 0.538 & $-$0.798 & -- \\
    3DGS-U~\cite{galappaththige2026predictive} & \second{0.313} & \second{$-$1.170} & \second{2.36} \\
    PRIMU~\cite{gottwald2026primu} & 0.356 & $-$0.871 & 3.31 \\
    \arrayrulecolor{black!25}\midrule\arrayrulecolor{black}
    \ours (ours) & \first{0.207} & \first{$-$1.472} & \first{1.52} \\
    \bottomrule
  \end{tabular}
\end{table}

%% file: tables/supp_runtime.tex
\begin{table}[t]
  \centering
  \small
  \setlength{\tabcolsep}{4pt}
  \caption{Runtime and peak GPU memory on one N3DV scene (cook\_spinach: 5,700 training images,
  300 test frames; backbone 4DGaussians) on one RTX 4090.
  Fit: one-off time per scene on the trained backbone, calibration included. Per frame: scoring one test frame.
  GPU: peak device memory, including the CUDA context.}
  \label{tab:supp_runtime}
  \begin{tabular}{@{}lccc@{}}
    \toprule
    & Fit (min) & Per frame (s) & GPU (GB) \\
    \midrule
    Backbone, training & 20.0 & -- & 7.6 \\
    \midrule
    FisherRF~\cite{jiang2024fisherrf} & 13.0 & 0.02 & 2.7 \\
    3DGS-U~\cite{galappaththige2026predictive} & 4.6 & 0.03 & 1.5 \\
    PRIMU~\cite{gottwald2026primu} & 7.0 & 0.60 & 2.4 \\
    \arrayrulecolor{black!25}\midrule\arrayrulecolor{black}
    \ours (ours) & 6.2 & 0.06 & 3.6 \\
    \bottomrule
  \end{tabular}
\end{table}